\documentclass[preprint,authoryear,3p,12pt]{elsarticle}

\usepackage{graphics}
\usepackage{epsfig}
\usepackage{graphicx}

\usepackage{amssymb}
\usepackage{amsmath}
\usepackage{subfigure}
\usepackage{color}
\usepackage{setspace}
\usepackage{rotating}
\usepackage{url}
\usepackage{gensymb}
\usepackage{floatrow}
\usepackage{multirow}
\usepackage{float}

\usepackage{array}
\newcolumntype{S}{>{\centering\arraybackslash}m{0.6cm}}
\newcolumntype{M}{>{\centering\arraybackslash}m{2.4cm}}
\newcolumntype{L}{>{\centering\arraybackslash}m{8.5cm}}

\usepackage{lineno}

\definecolor{darkgreen}{rgb}{0,.75,0}

\DeclareMathOperator*{\argmax}{arg\,max}

\newif\ifsqueezing
\newif\ifaggressivesqueezing

\squeezingtrue
\aggressivesqueezingfalse

\ifsqueezing
\ifaggressivesqueezing
\newcommand{\Caption}[1]{\caption{{\footnotesize #1}}}
\newcommand{\Section}[1]{\vspace{-3mm} \section{#1} \vspace{-2mm}}

\else
\newcommand{\Caption}[1]{\caption{#1}}
\newcommand{\Section}[1]{\vspace{0mm} \section{#1} \vspace{-1mm}}

\fi
\else
\newcommand{\Section}[1]{\section{#1}}

\newcommand{\Caption}[1]{\caption{#1}}

\fi

\usepackage{color}

\DeclareGraphicsExtensions{.png,.jpg}

\journal{ISPRS Journal of Photogrammetry and Remote Sensing~}

\begin{document}

\begin{frontmatter}



\title{From Google Maps to a Fine-Grained Catalog of Street trees}



\author[label1]{Steve~Branson\fnref{label3}}
\author[label2]{Jan~Dirk~Wegner\fnref{label3}}
\author[label1]{David~Hall}
\author[label2]{Nico~Lang}
\author[label2]{Konrad~Schindler}
\author[label1]{Pietro~Perona}
\address[label1]{Computational Vision Laboratory, California Institute of Technology, USA}
\address[label2]{Photogrammetry and Remote Sensing, ETH Z\"urich, Switzerland}
\fntext[label3]{joint first authorship}

\begin{abstract}
Up-to-date catalogs of the urban tree population are of importance for municipalities to monitor and improve quality of life in cities. Despite much research on automation of tree mapping, mainly relying on on dedicated airborne LiDAR or hyperspectral campaigns, tree detection and species recognition is still mostly done manually in practice.   
We present a fully automated tree detection and species recognition pipeline that can process thousands of trees within a few hours using publicly available aerial and street view images of Google Maps\texttrademark. These data provide rich information from different viewpoints and at different scales from global tree shapes to bark textures. Our work-flow is built around a supervised classification that automatically learns the most discriminative features from thousands of trees and corresponding, publicly available tree inventory data. In addition, we introduce a change tracker that recognizes changes of individual trees at city-scale, which is essential to keep an urban tree inventory up-to-date. The system takes street-level images of the same tree location at two different times and classifies the type of change (e.g., tree has been removed). Drawing on recent advances in computer vision and machine learning, we apply convolutional neural networks (CNN) for all classification tasks.
We propose the following pipeline: download all available panoramas and overhead images of an area of interest, detect trees per image and combine multi-view detections in a probabilistic framework, adding prior knowledge; recognize fine-grained species of detected trees. In a later, separate module, track trees over time, detect significant changes and classify the type of change.
We believe this is the first work to exploit publicly available image data for city-scale street tree detection, species recognition and change tracking, exhaustively over several square kilometers, respectively many thousands of trees.
Experiments in the city of Pasadena, California, USA show that we can detect $>70\%$ of the street trees, assign correct species to $>80\%$ for 40 different species, and correctly detect and classify changes in $>90\%$ of the cases.       
\end{abstract}


\end{frontmatter}

\section{Introduction}\label{sec:intro}


Urban forests in the USA alone contain around 3.8 billion trees \citep{nowak2002}. A relatively small but prominent element of the urban forest are street trees. Street trees grow along public streets and are managed by cities and counties. The most recent estimate is that there are 9.1 million trees lining the streets of California, about one street tree for every 3.4 people\footnote{A rough estimate for Europe is given in \citep{pauleit2005}. The number of people per street tree strongly varies across European cities between 10 to 48 inhabitants per street tree. However, it is unclear (and in fact unlikely) if US and European census numbers rely on the same definitions.} living in an urban area, with an estimated replacement value of \$2.5 billion \citep{mcpherson2016}. However, the greatest value of a street tree is not its replacement value but its ecosystem services value, i.e., all economic benefits that a tree provides for a community. These benefits include: a reduction in energy use, improvement in air and water quality, increased carbon capture and storage, increased property values and an improvement in individual and community well­being \citep{nowak2002,mcpherson2016}\footnote{The most recent estimate of the ecosystem services value of the street trees in California is \$1 billion per year or \$111 per tree, respectively \$29 per inhabitant.}. Still, inventories are often lacking or outdated, due to the cost of surveying and monitoring the trees.

We propose an automated, image-based system to build up-to-date tree inventories at large scale, using publicly available aerial images and panoramas at street-level. The system automatically detects trees from multiple views and recognizes their species. It draws on recent advances in machine learning and computer vision, in particular deep learning for object recognition~\citep{krizhevsky2012,szegedy2015,simonyan2015,girshickICCV15fastrcnn,shaoqing15fasterRcnn}, fine-grained object categorization~\citep{wah2011,angelova2013,branson2013,deng2013,duan2013,krause2014,zhangECCV2014}, and analysis of publicly available imagery at large scale~\citep{hays2008,agarwal2009,anguelov2010google,majdik2013mav,russakovsky2015}. The method is build around a supervised classification that uses deep convolutional neural networks (CNN) to learn tree identification and speies classification from existing inventories.

Our method is motivated by TreeMapLA\footnote{\url{https://www.opentreemap.org/latreemap/map/}}, which aims to build a publicly available tree inventory for the greater Los Angeles area. Its goal is to collect and combine already existing tree inventories acquired by professional arborists. In case no reasonably up-to-date data is available, which is often the case, a smartphone app is used to task users in a crowd-sourcing effort to fill in data gaps. Unfortunately only few people (so-called citizen scientists) participate. Only a small number of trees, e.g., $\approx 1000$ out of more than 80,000 in Pasadena, have been mapped within the last 3 years. And often entries are incomplete (e.g., missing species, trunk diameter) or inaccurate (e.g., GPS position grossly wrong). It turns out that determining a tree's species is often the hardest and most discouraging part for citizen scientists. The average person does not know many species of tree, and even with tree identification tools, the prospect of choosing one option among tens or even hundreds is daunting.

We propose to automate tree detection and species recognition with the help of publicly available street-level panoramas and very high-resolution aerial images. The hope is that such a system, which comes a\t virtually no cost and enables immediate inventory generation from scratch, will allow more cities to gain access to up-to-date tree inventories. This will help to ascertain the diversity of the urban forest by identifying tree species determinants of urban forest management (e.g., if a pest arrives, an entire street could potentially lose its trees). Another benefit to a homogeneous inventory across large urban areas would be to fill in the gaps between neighboring municipalities and different agencies, allowing for more holistic, larger-scale urban forest planning and management. Each city’s Tree Master Plan would no longer exist in a vacuum, but account for the fact that the urban forest, in larger metropolitan areas, spreads out across multiple cities and agencies. 

Our system works as follows: It first downloads all available aerial images and street view panoramas of a specified region from a repository, in our example implementation Google Maps. A tree detector that distinguishes trees from all other scene parts and a tree species classifier are separately trained on areas where ground truth is available. Often, a limited, but reasonably recent tree inventory does exist nearby or can be generated, which has similar scene layout and the same tree species. The trained detector predicts new trees in all available images, and the detector predictions are projected from image space to true geographic positions, where all individual detections are fused. We use a probabilistic conditional random field (CRF) formulation to combine all detector scores and add further (learned) priors to make results more robust against false detections. Finally, we recognize species for all detected trees. Moreover, we introduce a change classifier that compares images of individual trees acquired at two different points in time. This allows for automated updating of tree inventories.

\section{Related work}\label{sec:relwork}

There has been  steady flow of research into automated tree mapping over the last decades. A multitude of works exist and a full review is beyond the scope of this paper (e.g., see~\citep{larsen2011,kaartinen2012} for a detailed comparison of methods). 

{\bf Tree delineation in forests} is usually accomplished with airborne LiDAR data~\citep{reitberger2009,lahivaara2014,zhang2014} or a combination of LiDAR point clouds and aerial imagery~\citep{qin2014,paris2015}. LiDAR point clouds have the advantage of directly delivering height information, which is beneficial to tell apart single tree crowns in dense forests. On the downside, the acquisition of dense LiDAR point clouds requires dedicated, expensive flight campaigns. Alternatively, height information can be obtained through multi-view matching of high-resolution aerial images~\citep{hirschmugl2007} but is usually less accurate than LiDAR due to matching artefacts over forest.
%

Only few studies attempt segmentation of individual trees from a single aerial image.~\citet{lafarge2010} propose marked point processes (MPP) that fit circles to individual trees. This works quite well in planned plantations and forest stands with reasonably well-separated trees. However, MMPs are notoriously brittle and difficult to tune with inference methods like simulated annealing or reversible jump Markov Chain Monte Carlo, which are computationally expensive. Simpler approaches rely on template  matching, hierarchies of heuristic rules, or scale-space analysis (see~\citet{larsen2011} for a comparison).    
 
{\bf Tree detection in cities} has gained attention since the early 2000s. Early methods for single tree delineation in cities were inspired by scale-space theory (initially also developped for forests by~\citet{brandtberg1998}). A common strategy is to first segment data into homogeneous regions, respectively 3D clusters in point clouds, and then classify regions/clusters into tree or background, possibly followed by a refinement of the boundaries with predefined tree shape priors or active contours. For example,~\citet{straub2003} segments aerial images and height models into consistent regions at multiple scales, then performs refinement with active contours. 
Recent work in urban environments~\citep{lafarge2012} creates 3D city models from dense aerial LiDAR  point clouds, and reconstructs not only trees but also buildings and the ground surface. After an initial semantic segmentation with a breakline-preserving MRF, 3D templates consisting of a cylindrical trunk and an ellipsoidal crown are fitted to the data points. 
Similarly, tree trunks have been modeled as cylinders also at smaller scales but higher resolution, using LiDAR point clouds acquired either from UAVs~\citep{jaakkola2010} or from terrestrial mobile mapping vehicles~\citep{monnier2012}. 

We are aware of only one recent approach for urban tree detection that, like our method, needs neither need height information nor an infra-red channel.~\citet{yang2009} first roughly classify aerial RGB images with a CRF into tree candidate regions and background. Second, single tree templates are matched to candidate regions and, third, a hierarchical rule set greedily selects best matches while minimizing overlap of adjacent templates. 
%
%
This detection approach (tree species recognition is not addressed) is demonstrated on a limited data set and it remains unclear whether it will scale to entire cities with strongly varying tree shapes.

{\bf Tree species classification} from remote sensing data either uses multi-spectral aerial\linebreak~\citep{leckie2005,waser2011} or satellite images~\citep{pu2012}, hyperspectral data~\citep{clark2005,roth2015}, dense (full-waveform) LiDAR point clouds~\citep{brandtberg2007,yao2012}, or a combination of LiDAR and multispectral images~\citep{heikkinen2011,korpela2011,heinzel2012}. Methods that rely on full-waveform LiDAR data exploit species-specific waveforms due to specific penetration into the canopy, and thus different laser reflectance patterns, of different tree species; whereas hyperspectral data delivers species-specific spectral patterns.
Most works follow the standard classification pipeline: extract a  small set of texture and shape features from images and/or LiDAR data, and train a classifier (e.g., Linear Discriminant  Analysis, Support Vector Machines) to distinguish between a small number of species (3 in~\citep{leckie2005,heikkinen2011,korpela2011}, 4 in~\citep{heinzel2012}, 7 in~\citep{waser2011,pu2012}). 

%
Most state-of-the-art remote sensing pipelines (except~\citep{yang2009}) have in common that they need dedicated, expensive LiDAR, hyperspectral, or RGB-NIR imaging campaigns. They exploit the physical properties of these sensors like species-specific spectral signatures, height distributions, or LiDAR waveforms. As a consequence, methods are hard to generalize beyond a specific sensor configuration, data sets tend to be limited in size, and temporal coverage is sparse. Tree detection and species ground truth has to be annotated anew for each test site to train the classifier. It thus remains unclear if such methods can scale beyond small test sites to have practical impact at large scale (of course, sampling designs can be used, but these deliver only statistical information, not individual tree locations and types).

An alternative to remote sensing is in-situ interpretaton, or gathering of tree leafs, that are then matched to a reference database~\citep{du2007,kumar2012,mouine2013,goeau2013,goeau2014}. Anyone can recognize particular plant species with smart-phone apps like \emph{Pl@ntNet}~\citep{goeau2013,goeau2014} and \emph{Leafsnap}~\citep{kumar2012} that are primarily meant to educate users about plants. Experience with a similar app to populate the web-based tree catalog opentreemap have shown that it is difficult to collect a homogeneous and complete inventory with in situ measurements. Each tree must be visited by at least one person, which is very time consuming and expensive, and often the resulting data is incomplete (e.g., missing species), and also inaccurate when amateurs are employed.  

In this paper we advocate to solely rely on standard RGB imagery that is publicly available via online map services at world-wide scale (or from mapping agencies in some countries). We compensate the lack of pixel-wise spectral or height signatures with the power of deep CNNs that can learn discriminative texture patterns directly from large amounts of training data. We use publicly available tree inventories as ground truth to train our models.
Our method presented in this paper extends the approach originally presented in~\citet{wegner2016}, by integrating the detection and species recognition components into a single system. We further add a change tracking module, which utilizes a Siamese CNN architecture to compare individual tree states at two different times at city-scale. This makes it possible to recognize newly planted trees as well as removed trees, so as to update existing tree inventories. Additionally, we discuss results in depth and present a detailed analysis of failure cases. All steps of the pipeline are evaluated in detail on a large-scale dataset, and potential failure cases are thoroughly discussed.



\section{Detection of trees}\label{sec:detection}

We detect trees in all available aerial and panorama images to automatically generate a catalog of geographic locations, with corresponding species annotations. In this section we describe the detection component of the processing pipeline and our adaptations to include multiple views per object and map data\footnote{We refer the reader to~\ref{sec:projections} for details on projections between images and geographic coordinates.}.

We use the open source classification library Caffe\footnote{\url{http://caffe.berkeleyvision.org/}}~\citep{jia2014}. Several of the most recent and best-performing CNN models build on Caffe, including Faster R-CNN~\citep{shaoqing15fasterRcnn}, which forms the basis of our tree detector.
Faster R-CNN computes a set of region proposals and detection scores $R=\{(b_j,s_j)\}_{j=1}^{|R|}$ per test image $X$, where each $b_j=(x_j,y_j,w_j,h_j)$ is a bounding box and $s_j$ is the corresponding detection score over features extracted from that bounding box. A CNN is trained to both generate region proposals and computing detection scores for them. The method is a faster extension of the earlier R-CNN~\citep{girshick14CVPR} and Fast R-CNN~\citep{girshickICCV15fastrcnn} methods. Note that we start training the tree detector with a Faster R-CNN version pre-trained on Pascal VOC~\citep{everingham2010pascal}. 

A minor complication to using conventional object detection methods is that our target outputs and training annotations are geographic coordinates (latitude/longitude)--they are points rather than bounding boxes. A simple solution is to interpret boxes as regions of interest for feature extraction rather than as physical bounding boxes around an object.  At train time we can convert geographic coordinates to pixel coordinates using the appropriate projection function $\mathcal{P}_v(\ell,c)$ and create boxes with size inversely proportional to the distance of the object to the camera. At test time, we can convert the pixel location of the center of a bounding box back to geographic coordinates using $\mathcal{P}^{-1}_v(\ell',c)$. Doing so makes it possible to train single-image detectors. 

\subsection{Multi-view detection}\label{sec:detection-multiview}

Establishing sparse (let alone dense) point-to-point correspondences between aerial overhead imagery and panoramas taken from a street view perspective is a complicated wide-baseline matching problem. Although promising approaches for ground-to-aerial image matching (for buildings) have recetly emerged (e.g., \citep{shan2014,Lin2015}), finding correspondences for trees is an unsolved problem. We circumnavigate exact, point-wise correspondence search by combining tree detector outputs. 

\begin{figure*}[t]
\centering
\includegraphics[width=0.99\linewidth]{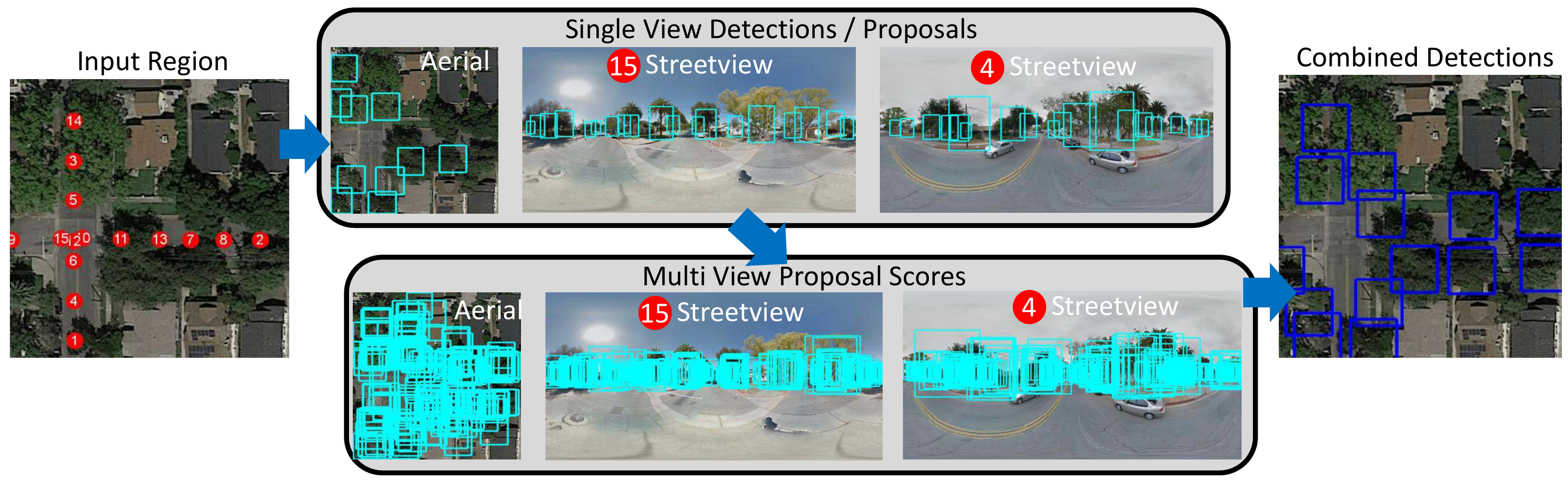}
\caption{\textbf{Multi View Detection:} We begin with an input region (left image), where red dots show available street view locations.  Per view detectors are run in each image (top middle), and converted to a common geographic coordinate system. The combined proposals are converted back into each view (bottom middle), such that we can compute detection scores with known alignment between each view. Multi-view scores are combined with semantic map data and spatial reasoning to generate combined detctions (right).  }
\label{fig:detection_flow}
\end{figure*}

For a given test image $X$, the algorithm produces a set of region proposals and detection scores $R=\{(b_j,s_j)\}_{j=1}^{|R|}$, where each $b_j=(x_j,y_j,w_j,h_j)$ is a bounding box and $s_j=\mathrm{CNN}(X,b_j;\gamma)$ is a corresponding detection score over CNN features extracted from image $X$ at location $b_j$. The region proposals can be understood as a short list of bounding boxes that might contain valid detections. In standard practice, a second stage is used where detection scores are thresholded and non-maximal suppression is used to remove overlapping detections. Naively performing this full detection pipeline can be problematic when combining multiple views (i.e., aerial and street views). Bounding box locations in one view are not directly comparable to another, and problems occur when an object is detected in one view but not the other. We do the following (Fig.~\ref{fig:detection_flow}):
\begin{enumerate}
\itemsep0em 
  \item Run the Faster R-CNN detector with a liberal detection threshold to compute an over-complete set of region proposals $R_v$ per view $v$. 
	\item Project detections of all views to geographic coordinates $R=\{\mathcal{P}_v^{-1}(\ell_{vj},c_v)\}_{j=1}^{|R_v|}$, with $\ell_{vj}$ the pixel location of the $j$-th region center.
  \item Collect all detections in geographic coordinates in the multi-view region proposal set $R$ by taking the union of all individual view proposals $R_v$.
	\item Project each detection region $\ell_k$ back into each view $v$ (with image coordinates $\mathcal{P}_v(\ell_k,c)$) 
	\item Evaluate all detection scores in each view $v$ of the combined multi-view proposal set $R$.
	\item Compute a combined detection score over all views, apply a detection threshold $\tau_2$ and suppress overlapping regions to obtain detections in geographic coordinates.
\end{enumerate}

This work-flow is robust to initially missing detections from single views because it collects all individual detections in geographic space and projects these back to all views, i.e. scores are evaluated also in views where nothing was detected in the first place. 
%

\subsection{Probabilistic model}\label{sec:detection-multiview-prob}

It seems inadequate to simply sum over detection scores per tree, because some views may provide more reliable information than others. In this section, we describe how we combine and weigh detections. The proposed probabilistic framework that is capable of also including prior information. 


Conditional Random Fields (CRF) provide a discriminative, probabilistic framework to meaningfully integrate different sources of information. They allow for construction of expressive prior terms (e.g.,~\citep{kohliIJCV2009,kraehenbuehl2011,wegner2015}) to model long-range contextual object-knowledge that cannot be learned from local pixel neighborhoods alone. Here, we formulate a CRF to combine tree detections from multiple views and to add further knowledge about typical distances from roads and spacing of adjacent trees.   
More formally, we aim at choosing the optimal set of tree objects $T$, based on evidence from aerial view imagery, street view imagery, semantic map data (e.g., the location of roads), and spatial context of neighboring objects (each $t_i \in T$ represents a metric object location in geographic coordinates):

\begin{equation}
\label{eq:detection_crf}
 \log p(T) = \sum_{t \in T} \biggl( \underbrace{k_{1}\Psi(t,\mathrm{av}(t);\gamma)}_{\mathrm{aerial\ view\ image}} + k_{2}\sum_{s \in \mathrm{sv}(t)} (\underbrace{\Phi(t,s;\delta)}_{\mathrm{street\ view\ images}}) + \underbrace{k_{3}\Lambda(t,T;\alpha)}_{\mathrm{spatial\ context}} + \underbrace{k_{4}\Omega(t,\mathrm{mv}(t);\beta)}_{\mathrm{map\ image}}\biggr) - Z
\end{equation}

where $\mathrm{lat}(t)$, $\mathrm{lng}(t)$ are shorthand for latitude and longitude of $t$; $\mathrm{av}(t)$, $\mathrm{mv}(t)$ are aerial and map view image IDs where tree $t$ is visible, and $\mathrm{sv}(t)$ is the street view image set ID that contains tree $t$ (with associated meta-data for the camera position). 
Potential functions $\Psi(\cdot)$ and $\Phi(\cdot)$ represent detection scores from aerial and street view images whereas $\Lambda(\cdot)$ and $\Omega(\cdot)$ encode prior knowledge. Parameters $\alpha$, $\beta$, $\delta$, $\gamma$ of the individual potentials are learned from training data whereas scalars $k_1$, $k_2$, $k_3$, $k_4$ that weight each potential term separately are trained on validation data (more details in~\ref{sec:trainInfer}). $Z$ is a normalization constant that turns overall scores into probabilities.  

The aerial view potential $\Psi(t,\mathrm{av}(t);\gamma)$ is the detection score evaluated at aerial image $X(\mathrm{av}(t))$:
\begin{equation}
 \Psi(t,\mathrm{av}(t);\gamma) = \mathrm{CNN}\left(\mathrm{X}(\mathrm{av}(t)),\mathcal{P}_{av}(t);\gamma\right)
\end{equation}
where $\mathcal{P}_{av}(t)$ transforms between pixel location in the image and geographic coordinates. $\gamma$ represents all weights of the aerial detection CNN learned from training data.

The street view potential $\Phi(t,s;\delta)$ for a street view image $X(s)$ is
\begin{equation}
 \Phi(t,s;\delta) = \mathrm{CNN}\left(\mathrm{X}(s),\mathcal{P}_{sv}(t,\mathrm{c}(s));\delta\right)
\end{equation}
where $\mathcal{P}_{sv}(t,c)$ (defined in Eq.~\ref{eq:streetview_geo2pix}) projects a pixel location in the image to geographic coordinates. We empirically found that simply taking the closest street view image per tree worked best to evaluate detections\footnote{Straight-forward combination of detections of the same tree in multiple street view images performed worse probably due to occlusions. A more elaborate approach would be to weight detections inversely to the distance between acquisition position and tree candidate, for example. We leave this for future work.}. $\delta$ encodes all weights of the street view detection CNN learned from training data.



The spatial context potential $\Lambda(t,T;\alpha)$ encodes prior knowledge about the spacing of adjacent trees along the road. Two different trees can hardly grow at exactly the same location and even extremely closely located trees are rather unlikely, because one would obstruct the other from the sunlight. The large majority of trees along roads have been artificially planted by city administration, which results in relatively regular tree intervals parallel to the road. We formulate this prior on spacing between trees as an additional potential, where the distribution of distances between neighboring objects is learned from the training data set: 
\begin{equation}
\Lambda(t,T;\alpha) = \alpha \cdot Q_s(d_s(t,T))
\label{eq:spatial_potential}
\end{equation}
where $d_s(t,T)=\min_{t' \in T}\|t-t'\|^2$ is the distance to the closest neighboring tree. $Q_s(d_s(t,T))$ is a quantized version of $d_s$, i.e. a vector in which each element is 1 if $d_s$ lies within a given distance range and 0 otherwise.  We then learn a vector of weights $\alpha$ (Fig.\ref{fig:spatial_context}(top)), where each element $\alpha_i$ can be interpreted as the likelihood that the closest object is within the appropriate distance range.


A second contextual prior $\Omega(t,\mathrm{mv}(t);\beta)$ based on map data models the common knowledge that trees rarely grow in the middle of a road but are usually planted alongside at a fixed distance. 
We compute distances for each pixel to the closest road based on downloaded maps\footnote{Roads in Google maps are white. Morphological opening removes other small symbols with grey-scale value 255, and morphological closing removes text written on roads.}. We formulate the spatial prior potential as
\begin{equation}
\Omega(t,\mathrm{mv}(t);\beta) = \beta \cdot Q_m(d_m(t))
\label{eq:map_potential}
\end{equation}
where $d_m(t)$ is the distance in meters between a tree $t$ and the closest road and, similar to the spatial context term $\Lambda(t,T;\alpha)$, function $Q_m()$ quantizes this distance into a histogram. Weight vector $\beta_i$ is learned from the training data set and entries can be viewed as likelihoods of tree-to-road distances (Fig.\ref{fig:spatial_context}(center)).

\subsection{Training and inference of the full model}\label{sec:trainInfer}

Inspired by~\citep{shotton2006textonboost} we use piecewise training to learn CRF parameters $\alpha^*,\beta^*,\delta^*,\gamma^*=\arg\max_{\alpha,\beta,\delta,\gamma} \log(p(T))$ that maximize Eq.~\ref{eq:detection_crf} (with $T$ the set of objects in our training set). 
%
%
The Pasadena data set is subdivided into train, validation, and test set. We first learn parameters for each potential term separately, optimizing conditional probabilities:
%
\begin{equation}
\begin{split} 
 \delta^* &= \argmax \sum_{t \in \mathcal{D}_t} \log p(t|\mathrm{av}(t))\\
 \log p(t|\mathrm{av}(t)) &= \Psi(t,\mathrm{av}(t);\gamma)-Z_3
\end{split} 
\label{eq:learn_aerial_view}
\end{equation}

\begin{equation}
\begin{split} 
 \gamma^* &= \argmax \sum_{t \in \mathcal{D}_t} \sum_{s \in \mathrm{sv}(t)} \log p(t|s)\\
 \log p(t|s) &= \Phi(t,s;\delta)-Z_4
\end{split} 
\label{eq:learn_street_view}
\end{equation}

\begin{equation}
\begin{split} 
 \alpha^* =& \argmax \sum_{t \in \mathcal{D}_t} \log p(t|T)\\
 \log p(t|T) =& \Lambda(t,T;\alpha)-Z_1
\end{split} 
\label{eq:learn_spatial_context}
\end{equation}

\begin{equation}
\begin{split} 
 \beta^* &= \argmax \sum_{t \in \mathcal{D}_t} \log p(t|\mathrm{mv}(t))\\
 \log p(t|\mathrm{mv}(t)) &= \Omega(t,\mathrm{mv}(t);\beta)-Z_2
\end{split} 
\label{eq:learn_map_context}
\end{equation}

where normalization terms $Z_{1...4}$ are computed for each training example individually to make probabilities sum to 1.  Note that the first two terms (Eq.~\ref{eq:learn_aerial_view} \& ~\ref{eq:learn_street_view}) match the learning problems used in Faster R-CNN training (which optimizes a log-logistic loss), and the last two terms are simple logistic regression problems (Eq.~\ref{eq:learn_spatial_context} \& ~\ref{eq:learn_map_context}).  

Next, we fix $\alpha,\beta,\delta,\gamma$ and use the validation set to learn scalars $k_1$, $k_2$, $k_3$, $k_4$ (Eq.~\ref{eq:detection_crf}) to weight each potential term separately. Here, we optimize detection loss (measured in terms of average precision) induced by our greedy inference algorithm. This allows us to learn a combination of the different sources of information while optimizing a discriminative loss. We iteratively select each scalar $k_i$ using brute force search.

\begin{figure}[t]
\centering
\includegraphics[width=.6\linewidth]{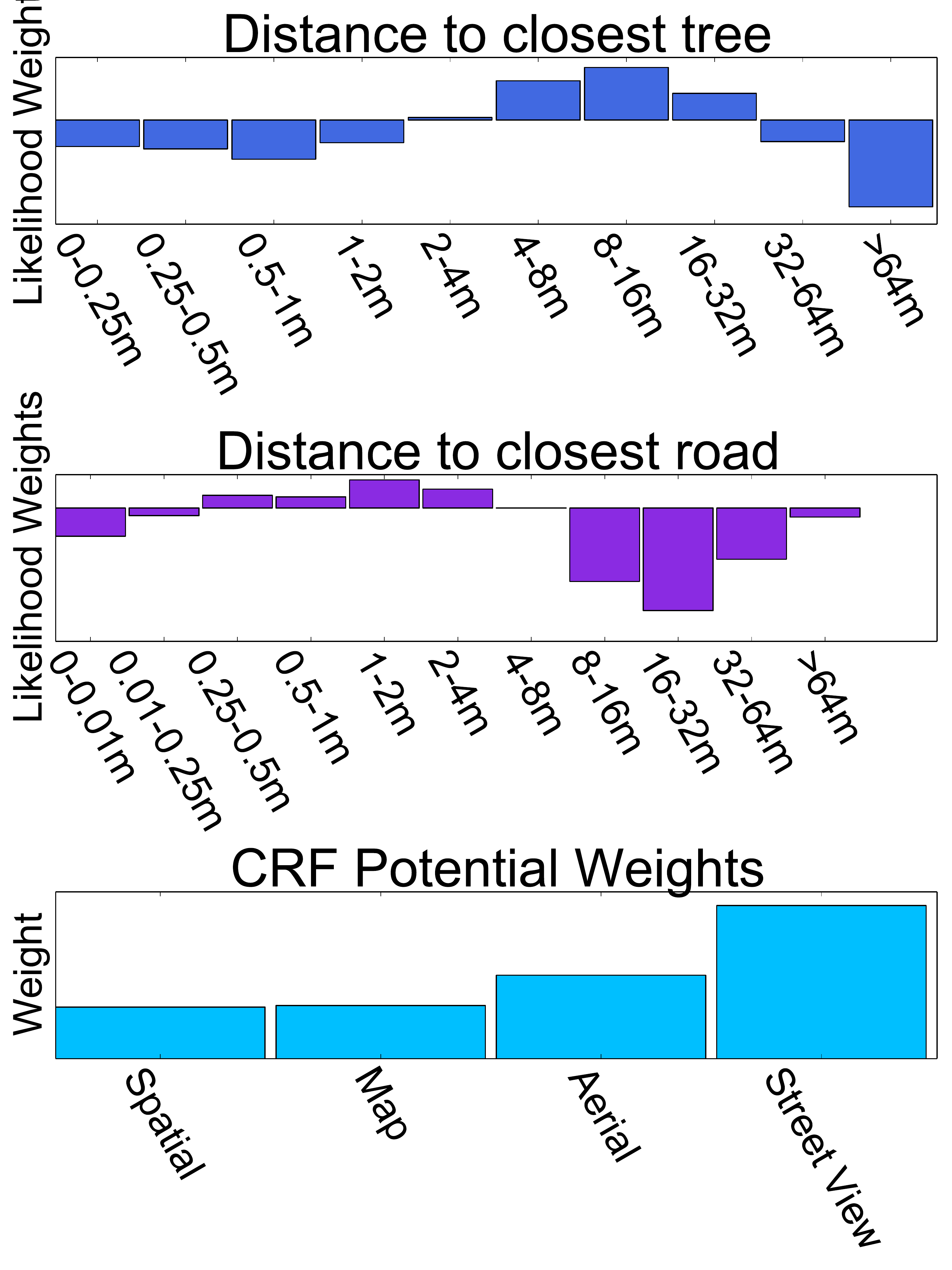}\\
\Caption{Visualization of learned spatial context parameters (top), map potential parameters (center), and (bottom) scalars $k_1$ (Aerial), $k_2$ (Street View), $k_3$ (Spatial), $k_4$ (Map) for combining detection CRF potentials (Eq.~\ref{eq:detection_crf}).}
\label{fig:spatial_context}
\end{figure}

In Figure~\ref{fig:spatial_context} we visualize components of the learned model. The first histogram shows learned weights $\alpha$ for the spatial context potential. Intuitively, we see that the model penalizes most strongly trees that are closer than 2m or further than 32m to the nearest tree. The 2$^\text{nd}$ histogram shows learned map weights $\beta$, the model penalizes trees that are too close ($<0.25$m) or too far ($>8$m) from the road. The last histogram shows learned weights $k_1$, $k_2$, $k_3$, $k_4$ on each CRF potential term, these match earlier results that streetview and aerial images are most important.

At test time the trained model is applied to before unseen, new data. The aim is to select the subset of tree detections $T^*=\arg\max_T \log(p(T))$ that maximize Eq.~\ref{eq:detection_crf}, which is generally a challenging (NP-hard) problem because all possible combinations would have to be tried. Here, we resort to a greedy approach that begins with $T=\emptyset$, and iteratively appends a new candidate detection
\begin{equation}
 t'=\argmax_t \log(p(T \cup t))
\end{equation}
until no new tree candidate remains that would increase $\log p(T)$. In fact, despite being a greedy procedure, this is quite efficient because we compute the combined detection score $\Omega(t,\mathrm{mv}(t);\beta)+\Psi(t,\mathrm{av}(t);\gamma) + \sum_{s \in \mathrm{sv}(t)} \Phi(t,s;\delta)$ only once for each location $t$ in the combined multi-view region proposal set $R$. That is, we can pre-compute three out of four potentials and only have to update the spatial term $\Lambda(t,T;\alpha)$ every time we add a new detection $t'$. Note that, in a probabilistic interpretation, the greedy accumulation is known to approximate non-maximum suppression~\citep{blaschkoICEMM11,qingNIPS15}, with known approximation guarantees for some choices of $\Lambda(\cdot)$.

\Section{Tree species classification}\label{sec:speciesrecognition}

Tree species recognition can be viewed as a n instance offine-grained object recognition, which has recently received a lot of attention in computer vision~\citep{wah2011,angelova2013,branson2013,deng2013,duan2013,krause2014,zhangECCV2014,russakovsky2015}. Fine-grained object recognition expands traditional object detectors (e.g., birds versus background) to distinguish subclasses of a particular object category (e.g., species of birds~\citep{branson2013}). In this sense, given detections of tree objects, our goal is to predict their fine-grained species. For this purpose we again use a CNN.

In contrast to our first work on this topic~\citep{wegner2016}, we no longer run two separate detection and species recognition modules\footnote{In~\citep{wegner2016} tree species recognition is demonstrated for ground truth tree locations.}, but directly recognize species at previously detected tree locations. More precisely, given tree detections we download one aerial image and three cropped regions at increasing zoom level of the closest street-view panorama for each tree. Examples of image sets for four different species of trees are shown in Fig.~\ref{fig:ImageExamples1}.
Here, we use the VGG16 network~\citep{simonyan2015} (in contrast to~\citep{wegner2016}, which used the shallower and less powerful GoogLeNet~\cite{szegedy2015}.


Four VGG16 CNNs are trained, separately per view and zoom level, to
extract features for each image. We follow the standard procedure and
deploy a log-logistic loss function and stochastic gradient descent
for training, where pre-trained model parameters on
ImageNet~\cite{russakovsky2015} are re-fined using our tree data
set. Fully-connected layers are discarded per CNN and features from all four views' networks are
concatenated into one feature vector per tree to train a linear SVM.
%

\Section{Tree change tracking}\label{sec:changetracking}

\begin{figure}[t]
\centering
\includegraphics[width=0.99\linewidth]{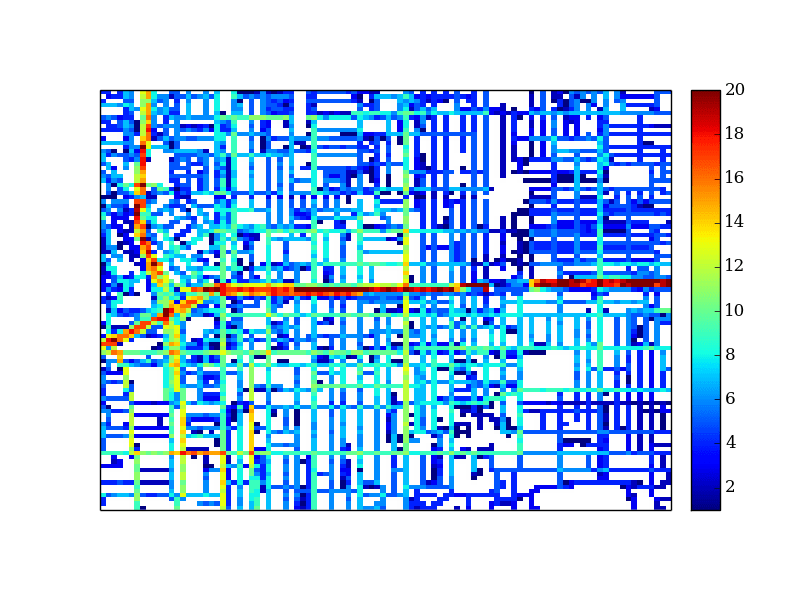}
\caption{Temporal street view coverage of Pasadena with number of images taken per $50m^{2}$ cell between 2006 and 2016. Empty cells are shown in white. }
\label{fig:sv_tempcov_pasadena}
\end{figure}

The simplest way for change tracking would be running the detection framework twice and comparing results. However, preliminary tests showed that much manual filtering was still necessary, for two main reasons: 
\begin{itemize}
\item Inaccuracies in the heading measurements of street-view panoramas and adding up of detection inaccuracies of two detector runs leads to the same tree often being mapped to two slightly different positions (by $\approx $1m - 8m). This in turn leads to many situations wrongly detected as changes. 
\item Revisit-frequency of the street-view mapping car is very inhomogeneous across the city. Main roads are mapped frequently whereas side roads have only very few images in total. Fig.~\ref{fig:sv_tempcov_pasadena} shows the revisit-frequency of the Google street view car in Pasadena as a heat map. As a consequence, it seems impossible to just run the detector for the entire scene for two different, exact points in time. A better strategy seems comparing per individual tree anytime new street-view data is acquired. 
\end{itemize}
We use another CNN for change tracking per individual tree, a so-called Siamese CNN. A Siamese architecture consists of two or more identical CNN branches that extract features separately from multiple input images. The branches share their weights, which implements the assumption that the inputs have identical statistics, and at the same time significantly reduces the number of free, learnable parameters.
\begin{figure}[!ht]
\centering
\includegraphics[width=.7\linewidth]{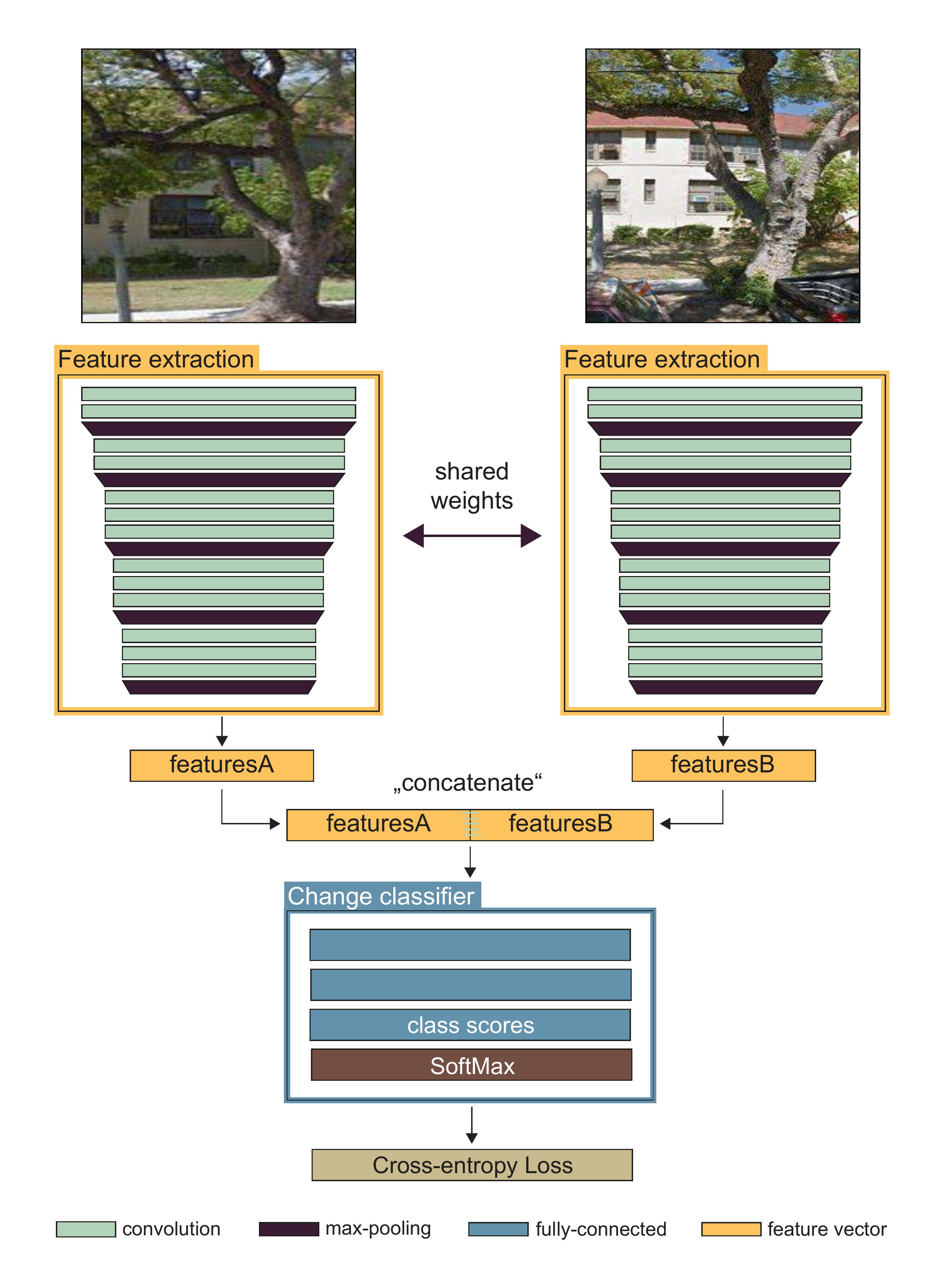}
\Caption{Siamese CNN network architecture.}
\label{fig:siameseDiagram}
\end{figure}
Two main strategies exist for similarity assessment via Siamese CNNs. The first passes extracted features to a contrastive loss~\citep{hadsell2006dimensionality}, which forces the network to learn a feature representation where similar samples are projected to close locations in feature space whereas samples that show significantly different appearance will be projected far apart (e.g.,~\citep{Lin2015,simo2015discriminative}. 
The second strategy uses Siamese CNNs as feature extraction branches that are followed by fully-connected layers to classify similarity (e.g.,~\citep{han2015matchnet,altwaijry2016learning}. We follow this change-classifier strategy because it best fits our problem of classifying change types of individual trees at different points in time. Network branches are based on the VGG-16 architecture ~\citep{simonyan2015}. We use VGG16 branches, because we can then start from our tree species classifier, which amounts to a pre-trained model already adapted to the visual appearance of street trees. Naturally, the top three fully-connected layers of the pre-trained VGG16 network are discarded, retaining only the convolutional and pooling layers as ``feature extractor''.
All features per network branch are concatenated and passed to the change classifier module, consisting of three fully-connected layers. The output of the top layer represents the class scores, which are passed through a SoftMax to get class probabilities. An illustration of our Siamese CNN architecture is shown in Figure \ref{fig:siameseDiagram}.
Starting from the pre-trained species classifier, the Siamese CNN is trained end-to-end for tree change classification. We use standard stochastic gradient descent in order to minimize the SoftMax cross-entropy loss. Batch-size is set to 16, implying that 16 image pairs are drawn per iteration, after having shuffled the data such that successive data samples within one batch never belong to a single class, so as to maximize the information gain per iteration.

Note that this first version of the tree change tracker needs given geographic positions to accurately compare per given tree position. It can already be useful in its present form if geographic coordinates per tree are available from GIS data. The siamese approach can then be directly applied.
%
%



\section{Experiments}\label{sec:experiments}

We evaluate our method on the Pasadena Urban Trees data set (Sec. \ref{sec:data}). First, we describe our evaluation strategy and performance measures. Then, we report tree detection, tree species classification, and tree change classification performance separately\footnote{Interactive demos of results for a part of Pasadena and the greater Los Angeles area ($>$1 million trees detected and species recognized) can be explored on our project website \url{http://www.vision.caltech.edu/registree/}.}. 

\subsection{Data and Test Site}\label{sec:data}

We densely download all relevant, publicly available aerial and street view images (and maps) from Google Maps 
within a given area. As test region we choose Pasadena, because a reasonably recent inventory with annotated species is available from 2013. Image data are from October 2014 (street view), respectively March 2015 (aerial images).
The Pasadena tree inventory is made available as kml-file that contains rich information 
(geographic position, genus, species, trunk diameter, street address, etc.) of $>140$ different tree species 
and about 80,000 trees in total in the city of Pasadena, California, USA. We found this to 
be a very valuable data source to serve as ground truth. The original inventory data set 
does not only contain trees but also potential planting sites, shrub etc. We filtered the 
data set such that only trees remain.
Densely downloading all images of Pasadena results in 46,321 street-level panoramas of size $1664\times832~px$ (and corresponding meta data and sensor locations), 28,678 aerial image tiles of size $256\times256~px$ (at $\approx 15~cm$ resolution), and 28,678 map tiles of size $256\times256~px$, over $>100,000$ images in total for our test region in the city of Pasadena. Panorama images are downloaded at medium resolution for the detection task, to speed up download and processing.   
A limitation of the Pasadena inventory is that it does only include trees on public ground, which we estimate constitute only $\approx 20\%$ of all trees in Pasadena. For training the tree detector we thus crowd-source labels for all trees (also those on private ground) in a subset of 1,000 aerial images and 1,000 street view panoramas, via Amazon Mechanical Turk\texttrademark. 
Note that we follow the standard approach and do zero-mean centering and normalization of all CNN input datasets (i.e., also for species recognition and change classification). More precisely, we first subtract the mean of all samples and normalize with the standard deviation of the training data. As a consequence, values of the pre-processed training image values range from 0 to 1 whereas values of validation and test data set may slightly exceed that range, but still have almost the same scale. 
%

\begin{figure}[H]
\centering
\begin{tabular}{cccc}
Aerial & SV40 & SV80 & SV110\\
\includegraphics[width=0.22\linewidth]{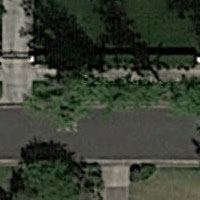} &
\includegraphics[width=0.22\linewidth]{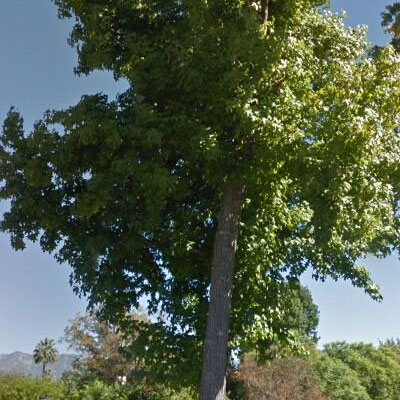} &
\includegraphics[width=0.22\linewidth]{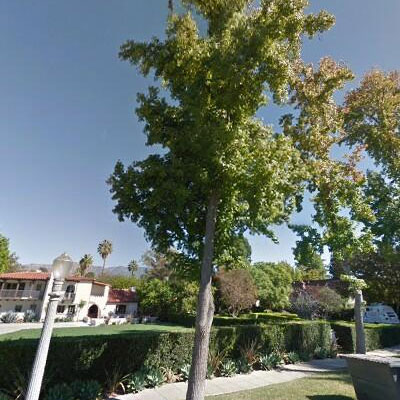} &
\includegraphics[width=0.22\linewidth]{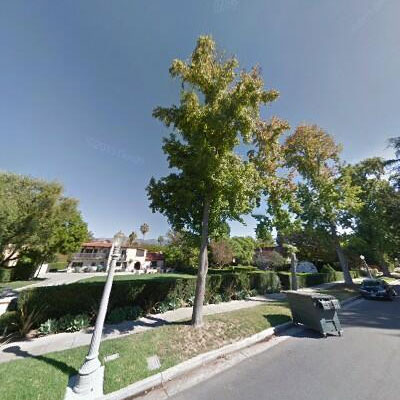} \\
(a) & (b) & (c) & (d) \\
%
\includegraphics[width=0.22\linewidth]{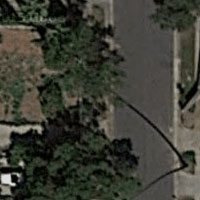} &
\includegraphics[width=0.22\linewidth]{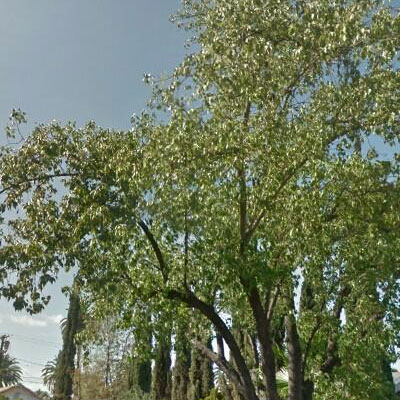} &
\includegraphics[width=0.22\linewidth]{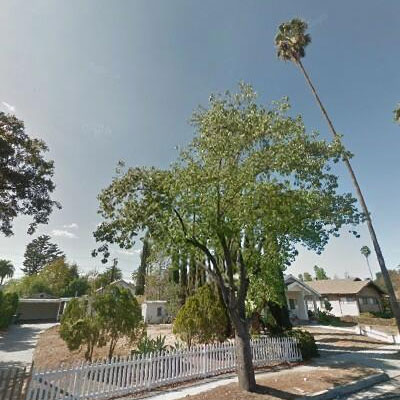} &
\includegraphics[width=0.22\linewidth]{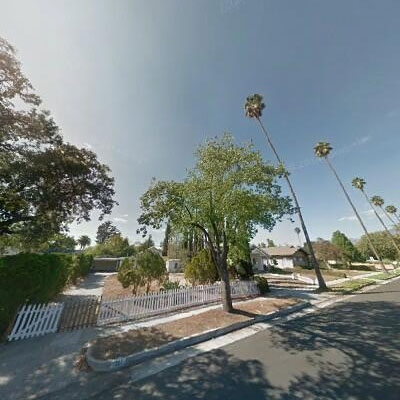} \\
(e) & (f) & (g) & (h) \\
%
\includegraphics[width=0.22\linewidth]{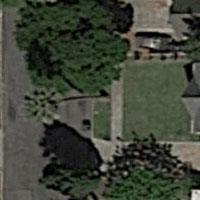} &
\includegraphics[width=0.22\linewidth]{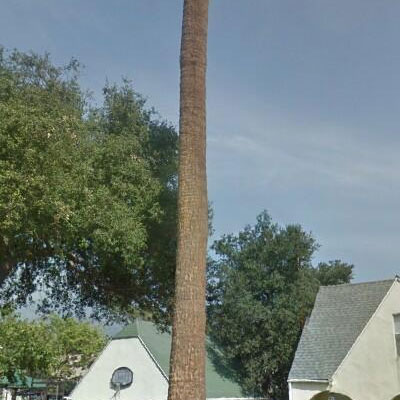} &
\includegraphics[width=0.22\linewidth]{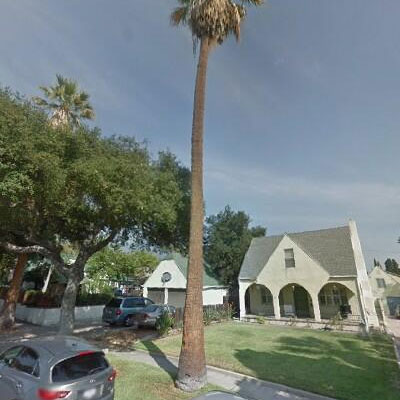} &
\includegraphics[width=0.22\linewidth]{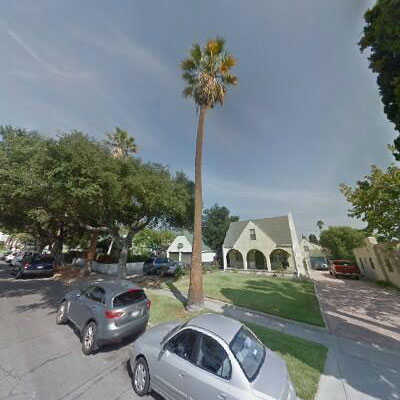} \\
(i) & (j) & (k) & (l) \\
%
\includegraphics[width=0.22\linewidth]{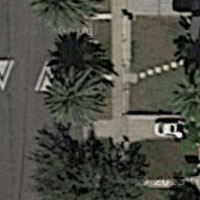} &
\includegraphics[width=0.22\linewidth]{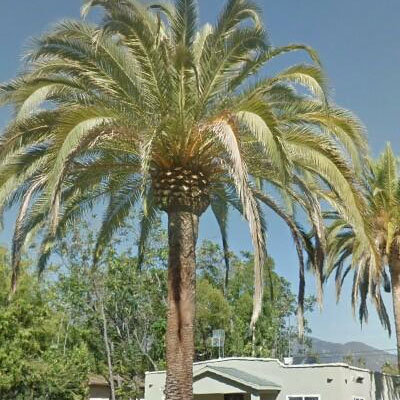} &
\includegraphics[width=0.22\linewidth]{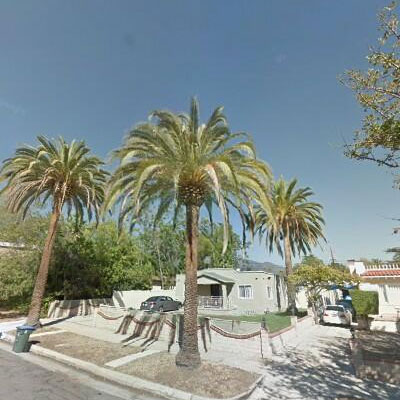} &
\includegraphics[width=0.22\linewidth]{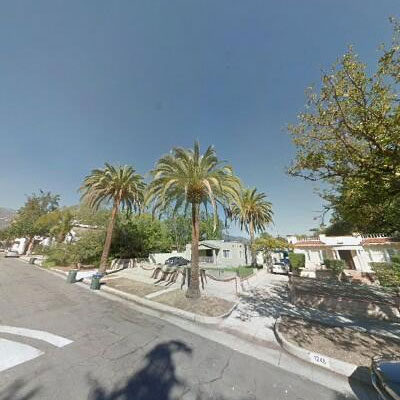} \\
(m) & (n) & (o) & (p) \\
\end{tabular}
\caption{Four images of the same tree (Row-wise top to bottom: American Sweet Gum, Bottle Tree, California Fan Palm, Canary Island Date Palm) automatically downloaded from Google servers if tasked with a specific geographic coordinate. From left to right: aerial image, zoom levels 40, 80, and 110 of the corresponding Google street view panorama.}
\label{fig:ImageExamples1}
\end{figure}

\begin{figure}[H]
\centering
\small
\begin{tabular}{cccc}
Mex. F. Palm: 2533 & Camph. Tree: 2250 & Live Oak: 1653 & Holly Oak: 1558\\
\includegraphics[width=0.22\linewidth]{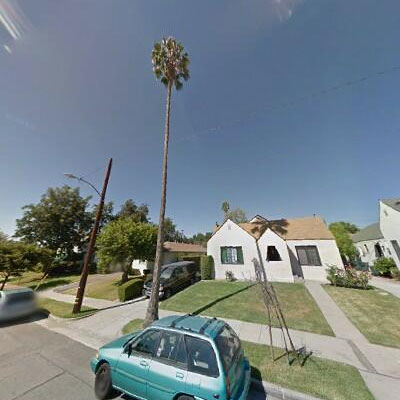} &
\includegraphics[width=0.22\linewidth]{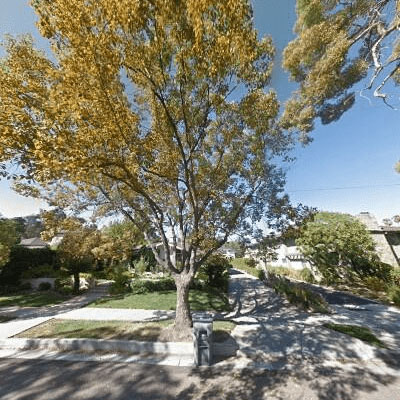} &
\includegraphics[width=0.22\linewidth]{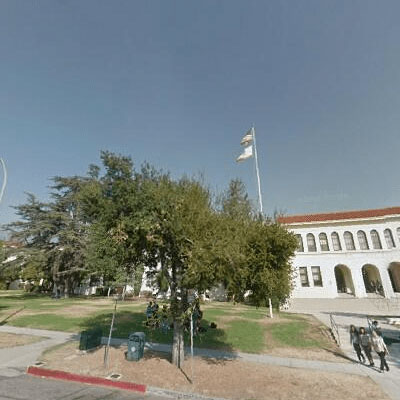} &
\includegraphics[width=0.22\linewidth]{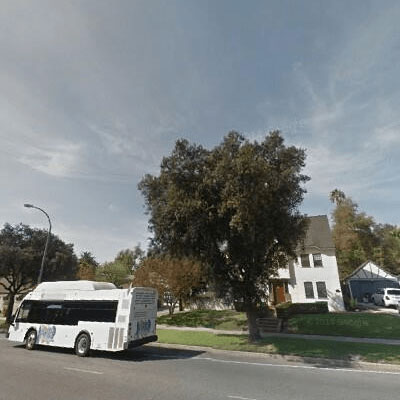} \\
South. Magn.: 1284 & Ca. I. D. Palm: 593 & Bottle Tree: 566 & Cal. Fan Palm: 522 \\
\includegraphics[width=0.22\linewidth]{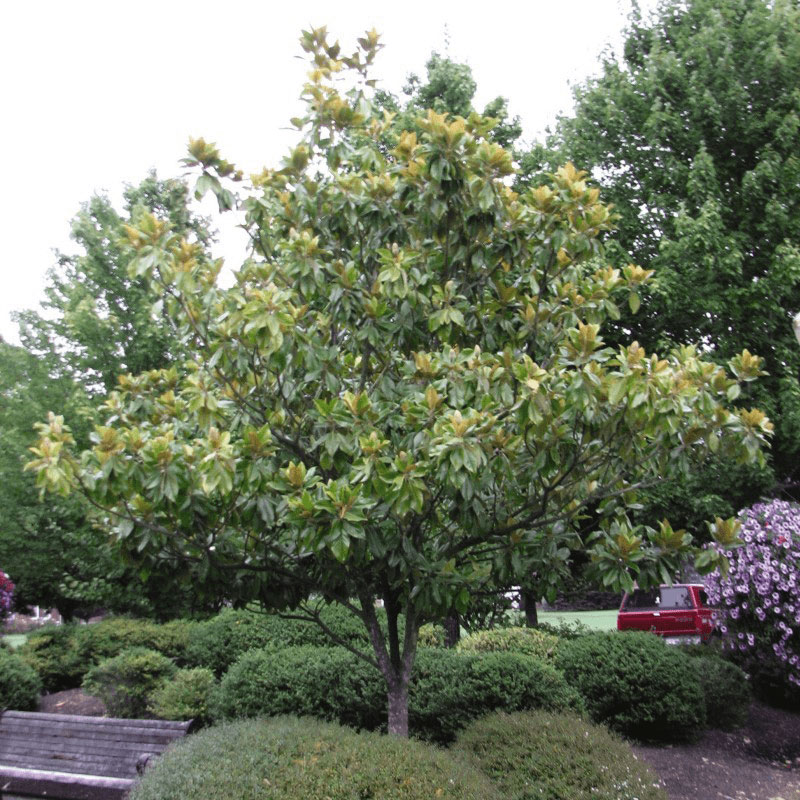} &
\includegraphics[width=0.22\linewidth]{Figures/CanaryIslandDatePalm_streetview_2043_fov_80_cropped_sorted} &
\includegraphics[width=0.22\linewidth]{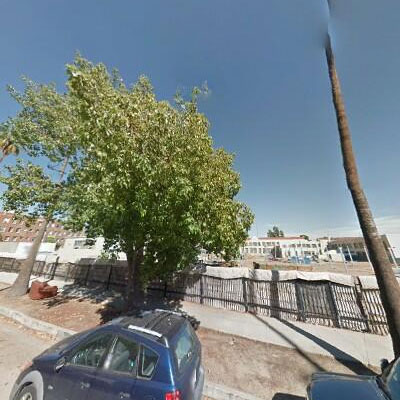} &
\includegraphics[width=0.22\linewidth]{Figures/CaliforniaFanPalm_streetview_2419_fov_110_cropped_sorted} \\
Ind. Laur. Fig: 335 & Chinese Elm: 330 & Jacaranda: 315 & Carob: 314 \\
\includegraphics[width=0.22\linewidth]{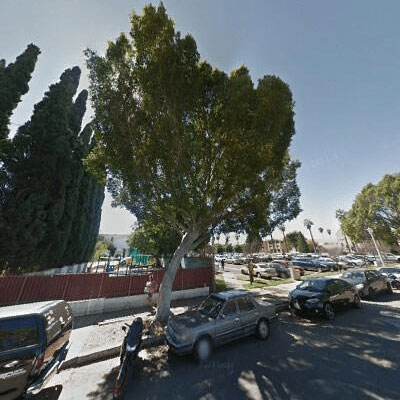} &
\includegraphics[width=0.22\linewidth]{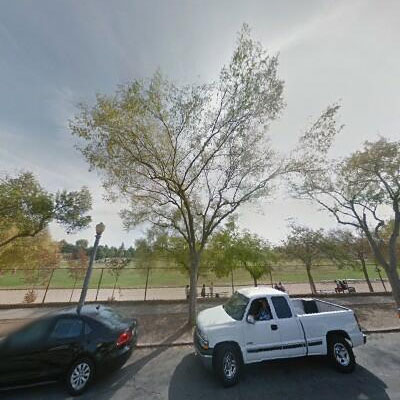} &
\includegraphics[width=0.22\linewidth]{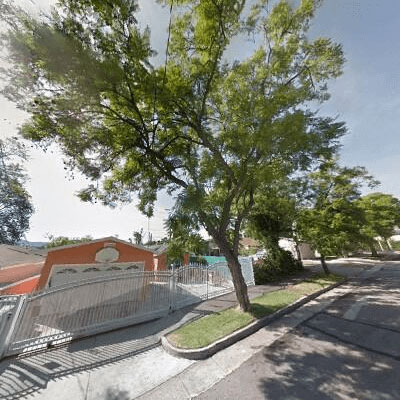} &
\includegraphics[width=0.22\linewidth]{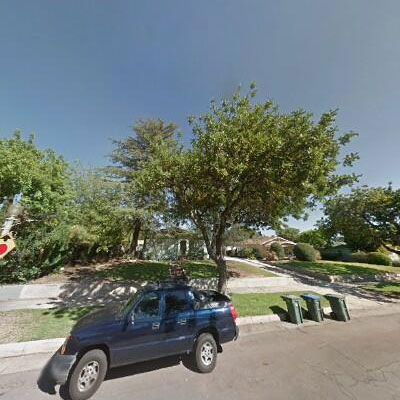} \\
Brush Cherry: 313 & Brisbane Box: 309 & Carrotwood: 305 & Italian Cypress: 270 \\
\includegraphics[width=0.22\linewidth]{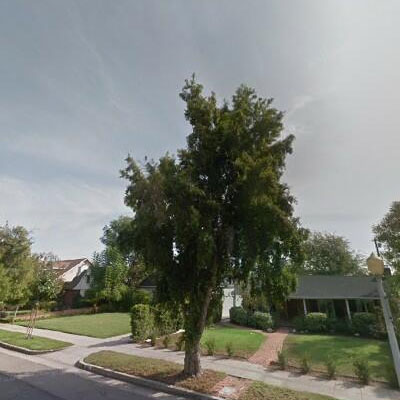} &
\includegraphics[width=0.22\linewidth]{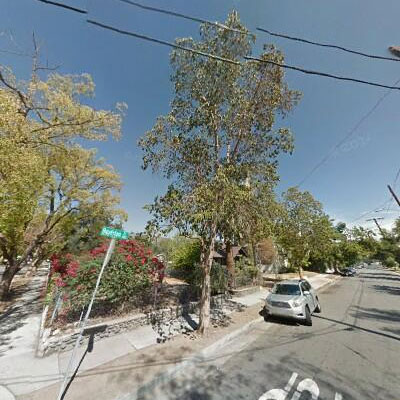} &
\includegraphics[width=0.22\linewidth]{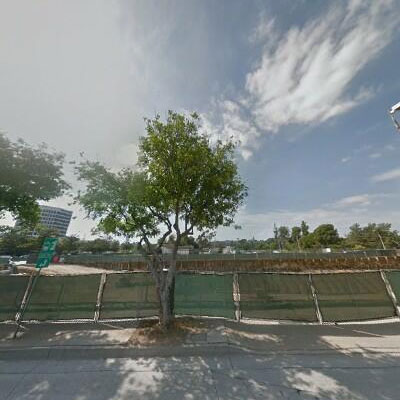} &
\includegraphics[width=0.22\linewidth]{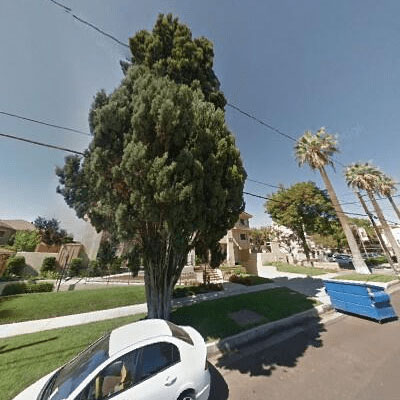} \\
Date Palm: 170 & Shamel Ash: 166 & Fern Pine: 160 & Am. Sweetgum: 155 \\
\includegraphics[width=0.22\linewidth]{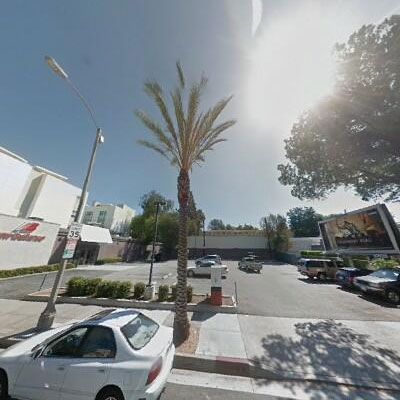} &
\includegraphics[width=0.22\linewidth]{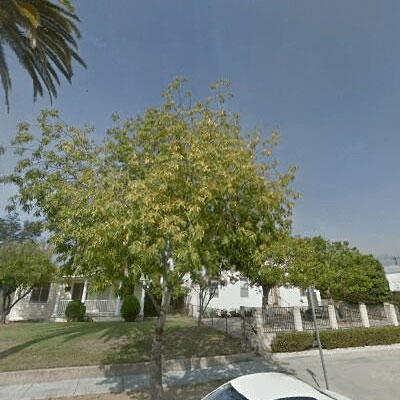} &
\includegraphics[width=0.22\linewidth]{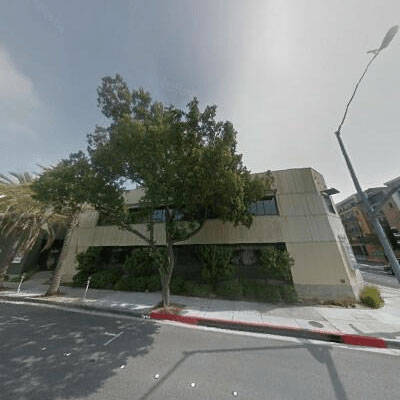} &
\includegraphics[width=0.22\linewidth]{Figures/AmericanSweetgum_streetview_18705_fov_110_cropped_sorted} \\
\end{tabular}
\caption{Example street-view images and total number of instances in our dataset for the 20 most frequent tree species. }
\label{fig:treeSpeciesNum}
\end{figure}

For species recognition of detected trees, we use four images per tree, one aerial 
image and three street-level images of different resolution (zoom) levels. If requesting street view data for a specific 
location, the Google server selects the closest panorama and automatically sets 
the heading towards the given location. Various parameters can be used to select 
particular images, the most important ones being the zoom level (i.e., level of 
detail versus area covered) and pitch. Preliminary tests showed that pitch could 
be fixed to 20 degrees, which leads to a slightly upward looking camera with 
respect to the horizontal plane (as measured during image acquisition by the 
Google camera car), which makes sense for flat ground and trees higher than the camera rig. In order to gather as much characteristic and discriminative 
appearance properties per tree as possible, street view imagery was downloaded at three 
different zoom levels (40, 80, 110) from the same point of view. This ensures that usually the entire tree (i.e., its 
overall, characteristic shape) as well as smaller details like branch structures, 
bark texture, and sometimes even leafs can be recognized. See example images 
for four different trees in Fig.~\ref{fig:ImageExamples1} and for the 20 most dominant species in Fig.~\ref{fig:treeSpeciesNum}. 

For tree change classification we download two images per tree acquired in 2011 and 2016, respectively. Both sets were processed with the detection algorithm. We then cut out an image patch at medium zoom level 80, as a compromise between level of detail and global tree shape, at each location in both image sets. We do this in both directions of time, i.e. we detect trees in the 2011 panoramas and extract image patches for 2011 and 2016 and we detect trees in the 2016 panoramas and extract patches for 2011 and 2016, suppressing double entries. This procedure ensures that we get image pairs showing both, removed and newly planted trees that show up at only one time, while also minimising the loss due to omission errors of the detector. We manually went through 11'000 image pairs to come up with a useful subset for experiments. Changes are relatively rare, the large majority of trees remains unchanged. From all manually checked image pairs we generate a balanced data subset of 479 image pairs with the three categories \textit{unchanged tree} (200 samples), \textit{new tree} (131), and \textit{removed tree} (148). 

\subsection{Evaluation strategy}\label{subsec:eval}

We randomly split the data into 70\% training, 20\% validation, and 10\% testing subsets for experiments. 

Evaluation of the tree detector uses precision and recall to assess detection performance by comparing results with ground truth. Precision is the fraction of all detected trees that match corresponding entries in ground truth. Recall refers to the fraction of ground truth trees detected by our method. Precision-recall curves are generated by sweeping through possible thresholds of the detection score, where each detection and ground truth tree could be matched at most once. More precisely, if two detected trees would equally well match the same ground truth tree, only one would be counted as true positive, the other as false positive. To summarize performance per tree detection variant in a single number, we report mean average precision (mAP) over all levels of recall\footnote{standard measure used in the VOC Pascal Detection Challenge~\citep{everingham2010pascal}}. 

We count a tree detection within a 4 meters radius of ground truth as true positive. This may seem high, but discussions with arborists showed that many tree inventories today, at least in the US, do not come with geographic coordinates but with less accurate street addresses. The major requirement in terms of positioning accuracy is that two neighboring trees can be distinguished in the field.
In these circumstances a 4 meter buffer seemed the most reasonable choice.
Empirically, most of our detections ($\approx 70\%$) are within $2~m$ of the ground truth, in part due to limited GPS and geo-coding accuracy.

For species classification we report dataset precision and average class precision. Dataset precision measures the global hit rate across the entire test set regardless of the number of instances per tree species. Consequently, those species that occur most frequently dominate this measure. In contrast, average class precision first evaluates precision separately per species and then returns the average of all separate values, i.e. it accounts for the long-tailed distribution of tree instances per species.
For tree change classification experiments we report full confusion matrices as well as mean overall accuracy (mOA).

\subsection{Tree detection results}\label{subsec:results_detection}

\begin{figure}[!ht]
\centering
\includegraphics[width=0.8\linewidth]{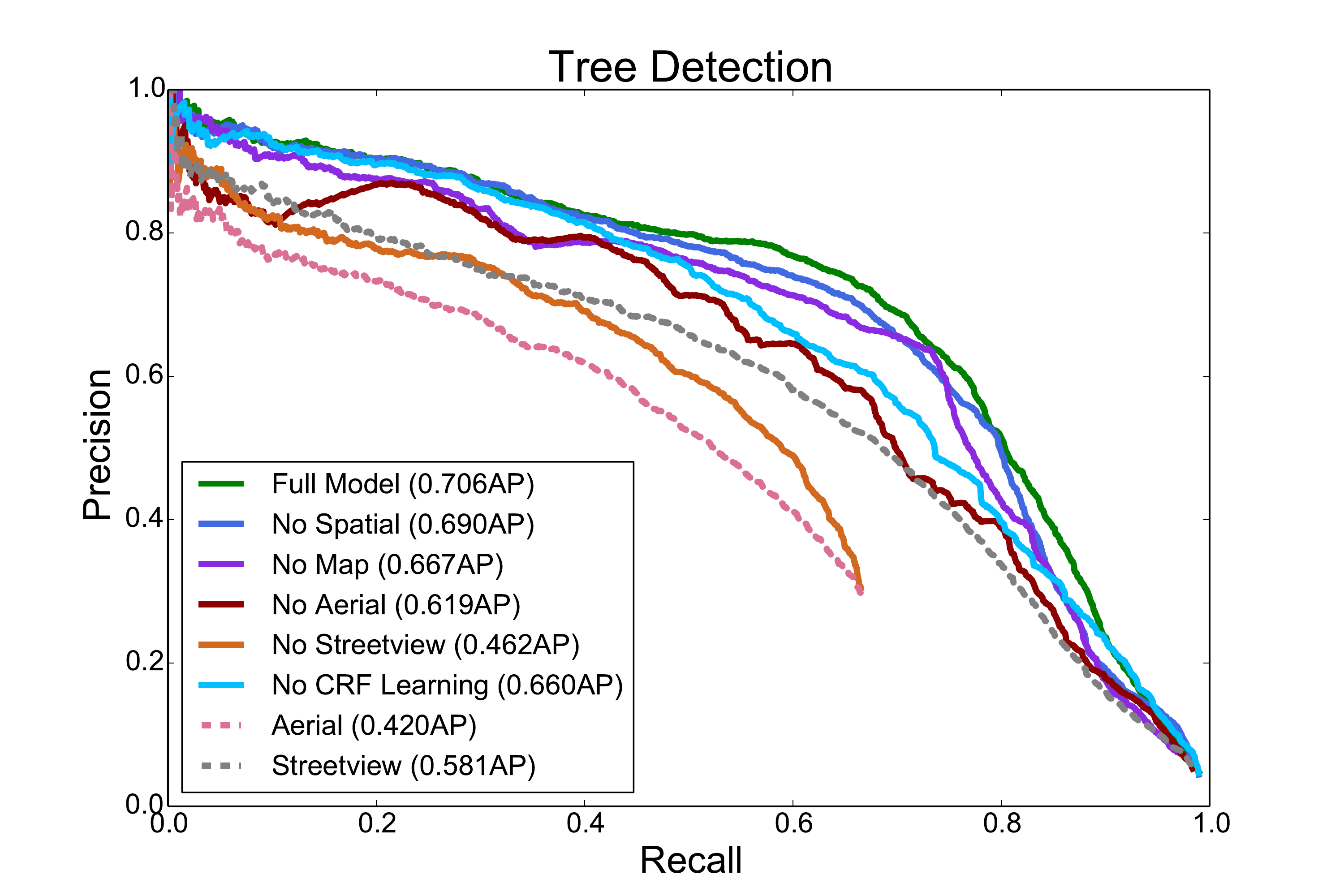} \\
\Caption{Precision-recall curve of the full tree detection model (green) compared to reduced models that are missing one of the prior terms and to only using aerial images (dotted rose) or only street level panoramas (dotted grey).}
\label{fig:detection_results}
\end{figure}

We plot precision-recall curves of our results in Figure~\ref{fig:detection_results}. It turns out that the combination of multiple views significantly improves results. A pure aerial image detector that applies the Faster R-CNN detection system~\citep{shaoqing15fasterRcnn} achieves only $0.42$ mAP (dotted rose curve in Fig. \ref{fig:detection_results}) compared to $0.706$ mAP of our full model (green curve in Fig. \ref{fig:detection_results}). Visual inspection of detections reveals that aerial images are very useful for identifying trees, but less so for accurately determining the location of the (occluded) trunks. Detections from aerial images can at best assume that a tree trunk is situated below the crown center, which might often not be the case. Particularly in cases of dense canopy, where individual trees can hardly be told apart from overhead views alone, this implicit assumption becomes invalid and results deteriorate. 
This situation changes if using street level panoramas because tree trunks can directly be seen in the images. A pure street view detector (dotted gray curve in Fig. \ref{fig:detection_results}) that combines multiple street views achieves better performance ($0.581$ mAP) than the aerial detector. 
This baseline omits all non-street view terms from Eq.~\ref{eq:detection_crf} and replaces the spatial context potential (Eq. \ref{eq:spatial_potential}) with a much simpler local non-maximum suppression that forbids neighboring objects to be closer than $\tau_{nms}$
\begin{equation}
\Lambda_{nms}(t,T;\alpha) = \begin{cases}
-\infty &\text{if } d_s(t,T) < \tau_{nms}\\
0 &\text{otherwise}
\end{cases}
\label{eq:spatial_potential_nms}
\end{equation}
In contrast to this simpler approach with a hard threshold, the learned approach has the advantage that it softly penalizes trees from being too close. It also learns that it is highly unlikely that a certain tree is completely isolated from all other trees.

We also find that in many cases detections evaluated as false positives are in fact trees, but located on private land near the road. The tree inventory of public street trees that we use as ground truth does not include these trees, which consequently count as false positives if detected by our system. Adding a GIS layer to the system, which specifies public/private land boundaries would help discarding tree detections on private land from evaluation. We view this as an important next step, and estimate that performance of the full model would increase by at least $5$\% in mAP.

To verify the impact of each potential term on detection results, we run lesion studies where individual parts of the full model are left away. Results of this experiment are shown in Figure~\ref{fig:detection_results}. "No Aerial", "No Streetview", and "No Map" remove the corresponding potential term from the full model in Eq.~\ref{eq:detection_crf}. "No Spatial" replaces the learned spatial context term (Eq.~\ref{eq:spatial_potential}) with a more conventional non-maximal suppression term (Eq.~\ref{eq:spatial_potential_nms}) introduced earlier in this section. 
We see the biggest loss in performance if we drop street view images ($0.706 \to 0.462$ mAP) or aerial images ($0.706 \to 0.619$ mAP). Dropping the map term results in a smaller drop in performance ($0.706 \to 0.667$ mAP). Replacing the learned spatial context potential with non-maximal suppression results in only a small drop ($0.706 \to 0.69$ mAP). For each lesioned version of the system we re-learn an appropriate weight for each potential function on the validation set. The method "No CRF Learning" shows results if we use the full model but omitted learning these scaling factors and set them all to 1 (results in a $0.706 \to 0.66$ mAP drop). 

\subsubsection{Detection Error Analysis}\label{subsubsec:detection_err_analysis}

\begin{figure}[!ht]
\centering\includegraphics[width=1.0\linewidth]{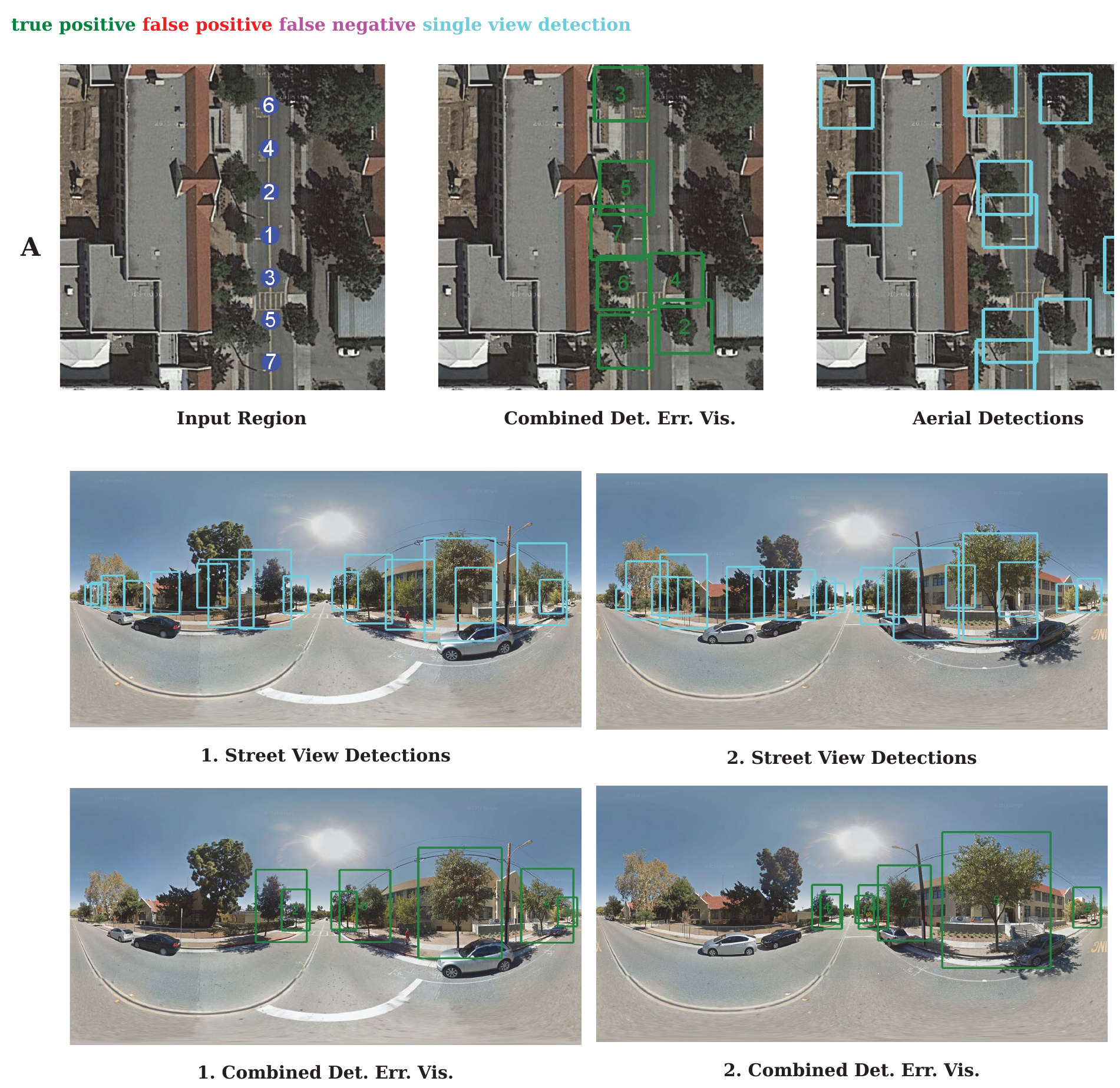}
\Caption{\textbf{Example A:} The detection system has correctly detected 7 trees, with no false positives or negatives.  It has also correctly rejected two large trees (top right of the input region) that are just on private property. }
\label{fig:detection_example_A}
\end{figure}

\begin{figure}[!ht]
\centering\includegraphics[width=1.0\linewidth]{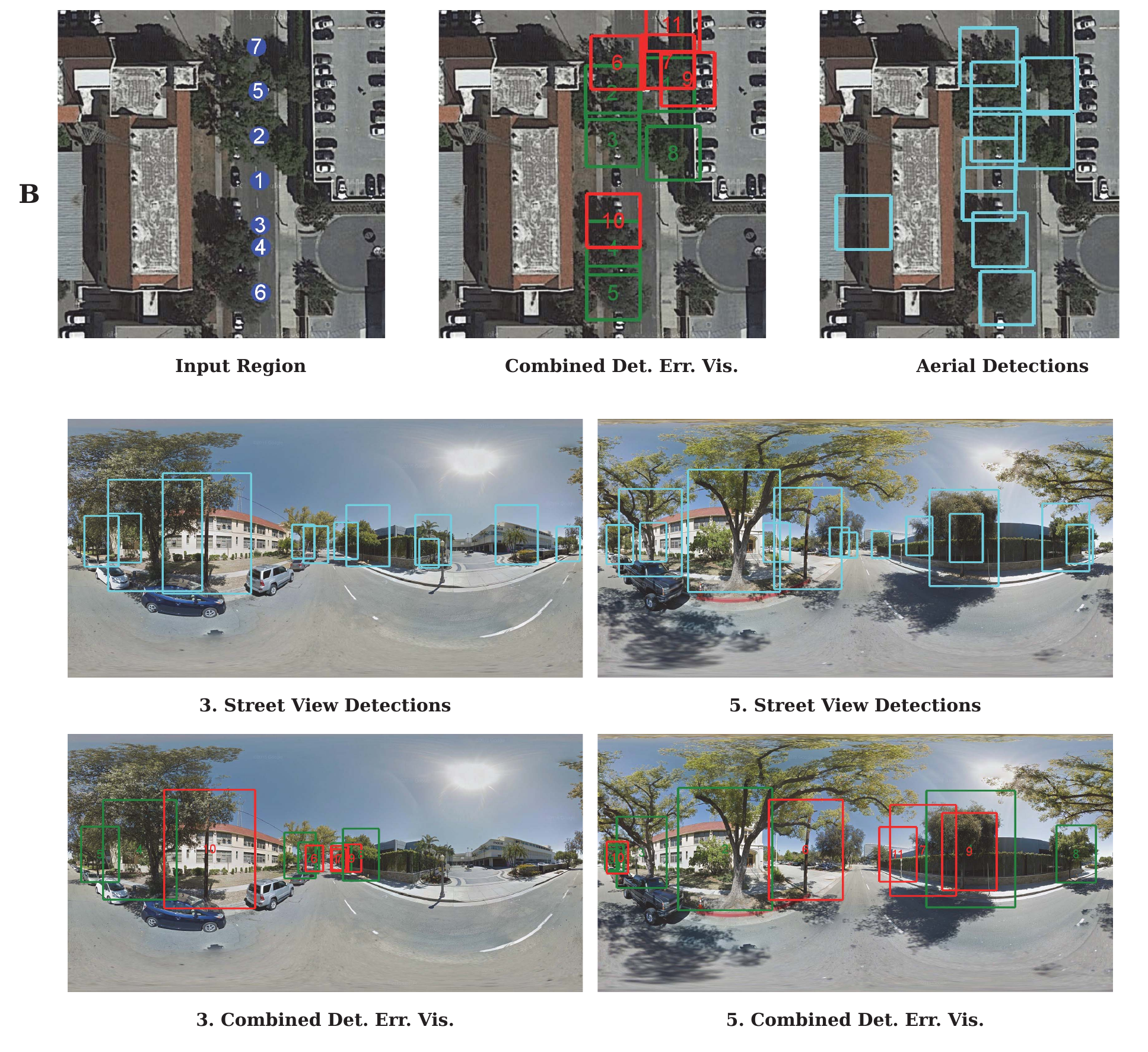}
\Caption{\textbf{Example B:} The detection system has correctly detected 6 trees, and correctly rejected a couple of trees on private property.  However, it has 5 false positives, including 3 false positives caused by wooden telephone poles near foliage (boxes 6, 10, 11), and 2 trees that were split into duplicate detections.}
\label{fig:detection_example_B}
\end{figure}

\begin{figure}[!ht]
\centering\includegraphics[width=1.0\linewidth]{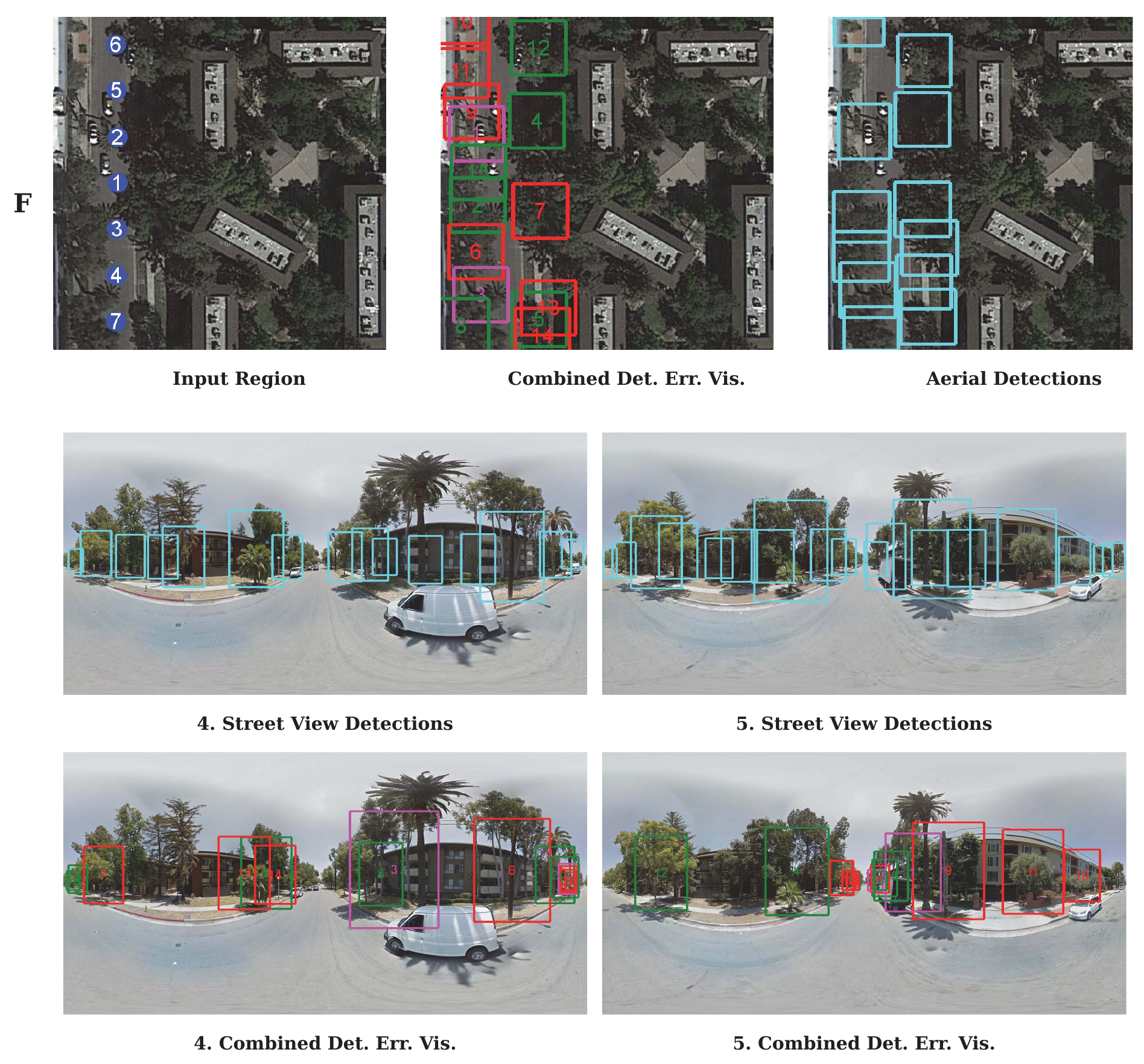}
\Caption{\textbf{Example F:} The detector has correctly detected 6 trees and correctly rejected upwards of 10 trees on private land.  However, there were 2 measured false positives (boxes 10, 11) that were actual trees but not included in the test set inventory because they are on private land.  A 3rd measured false positive appears to be a valid tree on public land, but was missing from the test set inventory for unknown reason.  3 other false positives occurred due to wooden telephone poles near foliage.  One false negative occurred due to a weak detection with score that fell just below the detection threshold (boxes 1), as it was detected in the street view image.  A second false negative probably occurred because a white van occluded the base of the trunk.}
\label{fig:detection_example_F}
\end{figure}

\begin{figure}[!ht]
\centering\includegraphics[width=1.0\linewidth]{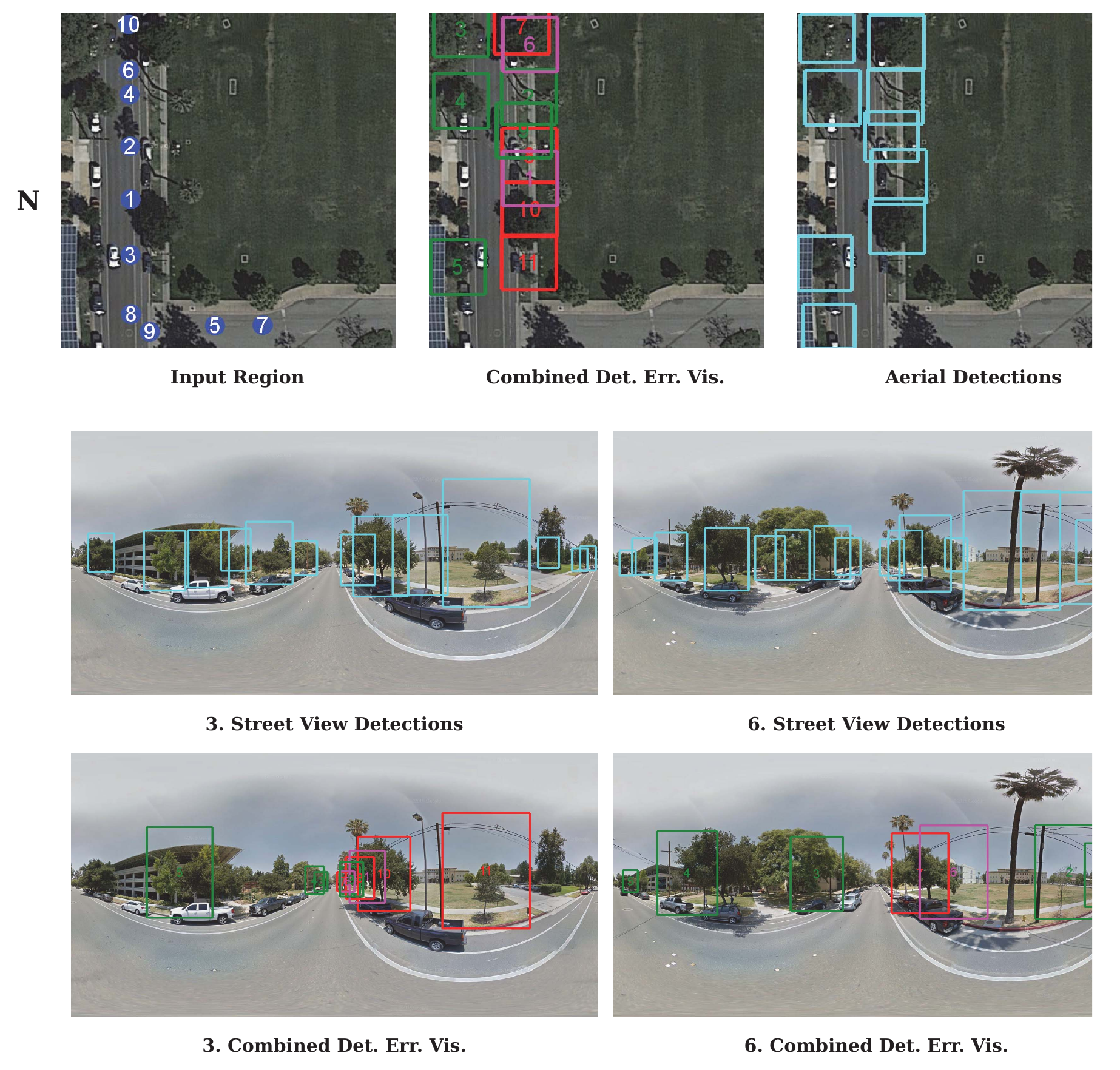}
\Caption{\textbf{Example N:} The detector correctly detected 5 trees and correctly rejected 5 trees on private land.  The detector also correctly detected 2 more trees (boxes 10,11) that were penalized as false positives because they were missing from the test set inventory.  They were probably missing because they appear to be recently planted and the inventory was collected in 2013.  The detector correctly detected 2 other trees; however, the localization was inaccurate, resulting in false positives (boxes 7,8) with respective false negatives (boxes 6,1).  The inaccuracy probably occurred because the recorded camera position and heading of the Google street view camera was noisy, as evidenced by the fact that boxes 7 and 8 are in the correct locations in the street view images but not the aerial image.  This is a common problem and subject of future research.}
\label{fig:detection_example_N}
\end{figure}

\begin{table}[!ht]
\begin{center}
\begin{tabular}{|M|L|S|S|M|}
\hline
  \textbf{Error Name} & \textbf{Error Description} & \textbf{\#} & \textbf{\%} & \textbf{Detection examples} \\
\hline
  \textbf{Private tree} & Detection corresponds to a real tree. The tree is on private land, whereas the inventory only includes trees on public land, resulting in a false positive. & 56 & 10.8 & {\color{red}F10, F11} \\
\hline
  \textbf{Missing tree} & A tree on public land appears to be missing from the inventory, which is older than the Google imagery (results in a false positive). Usually a recently planted tree. & 39 & 7.5 & {\color{red} F7, N10, N11} \\
\hline
  \textbf{Extra tree} & An extra  tree appears in the inventory. Often the case if a tree has been cut down since the inventory. & 66 & 12.7 & -\\
\hline
  \textbf{Telephone pole} & False positive because a telephone pole or lamp post resembles the trunk of a tree. This usually happens when foliage of a nearby tree also appears near the pole. & 49 & 9.4 & {\color{red}B6, B10, B11, F6, F9, F14}  \\
\hline
  \textbf{Duplicate detection} & A single tree is detected as 2 trees. & 19 & 3.6 & {\color{red}B7, B9, F13}  \\
\hline
  \textbf{Localization Error} & A detected tree is close to ground truth, but not within the necessary distance threshold, resulting in a false positive and negative. This usually happens when the camera position and heading associated with a Google street view panorama are slightly noisy. Another reason are inaccurate GPS positions in the inventory. & 40 & 7.7  & {\color{red}N7}/{\color{magenta}N6}, {\color{red}N8}/{\color{magenta}N1}  \\
\hline
  \textbf{Occluding object} & A tree is occluded (e.g., by a car or truck) in  street view, resulting in a false positive or error localizing the base of the trunk. & 120 & 23.1 & {\color{magenta} F3}  \\
\hline
  \textbf{Other false negatives} & An existing and clearly visible tree (included in the inventory) remains undetected. Often happens for very small, recently planted trees. Another reason is if the closest street view panorama is rather far away from a tree, which then appears small and blurry at low resolution in the images. & 131 & 25.2 & - 
\\
\hline
\end{tabular}
\end{center}
\caption{Analysis of a subset of 520 detection errors. Detection examples are shown overlaid to images in Figures~\ref{fig:detection_example_A}-\ref{fig:detection_example_N}.}
\label{tab:error_analysis}
\end{table}
In this subsection we present a detailed analysis of the full detection system. We manually inspect a subset of 520 error cases to analyze the most frequent failure reasons (see summary in Tab.~\ref{tab:error_analysis}). Examples for several typical failure cases are shown in Figures~\ref{fig:detection_example_A}-\ref{fig:detection_example_N}. 
In the top row of each figure, the first column shows the input region, with blue circles representing the location of available street view cameras. The 2nd column shows results and error analysis of our full detection system, with combined aerial, street view, and map images and spatial context.  Here, true positives are shown in green, false positives are shown in red, and false negatives are shown in magenta. The 3rd column shows single view detection results using just aerial images. The bottom two rows show two selected street view images--the images are numbered according to their corresponding blue circle in the 1st row, 1st column. The 2nd row shows single view detection results using just street view images. The bottom row visualizes the same results and error analysis visualized in the 1st row, 2nd column, with numbers in the center of each box matching across views.
We identify 8 main categories that can explain all 520 errors, and manually assign each error to one of the categories. Table~\ref{tab:error_analysis} details failure reasons, total numbers of errors per category, as well as percentage out of the sample of 520 investigated cases in total, and references to examples in Figures~\ref{fig:detection_example_A}-\ref{fig:detection_example_N}. We also include a likely explanation for each failure case in the image captions. 
Each example is denoted by a number and a letter, where the letter denotes the figure number, and the number denotes the bounding box number.  For example \textit{B3} corresponds to bounding box 3 (see numbered boxes in Figure~\ref{fig:detection_example_B}) in example B (Figure~\ref{fig:detection_example_B}).

We note that at least $31\%$ of measured errors (Private Tree, Missing Tree, Extra Tree) arise due to issues with the ground truth Pasadena inventory test set--these were correct detections that were penalized as false positives because the data set does not include trees on private land or because the inventory is less recent than Google maps imagery. Additionally, 40 ($7.7\%$) errors occurred due to detections that were close to ground truth trees, but localization was not quite accurate enough to meet the requisite distance threshold. Besides inaccuracies in aerial and street view geo-referencing and detections that can lead to this failure case, it should be noted that ground truth GPS positions are also off by several meters in some cases. Occlusions (mostly by trucks and cars) are the largest single error source ($23.1\%$) that causes trees to remain undetected or to be projected wrongly into geographic space.

\subsection{Tree species classification results}\label{subsec:results_classification}

Preliminary tests showed that we gain 5--10\% tree species recognition improvement if using VGG-16 models that have been  pre-trained on the ImageNet data set~\citep{russakovsky2015}. We thus start with pre-trained models for all experiments and fine-tune them on our data. 
\begin{figure}[!ht]
\centering
\begin{tabular}{cc}
\includegraphics[width=.5\linewidth]{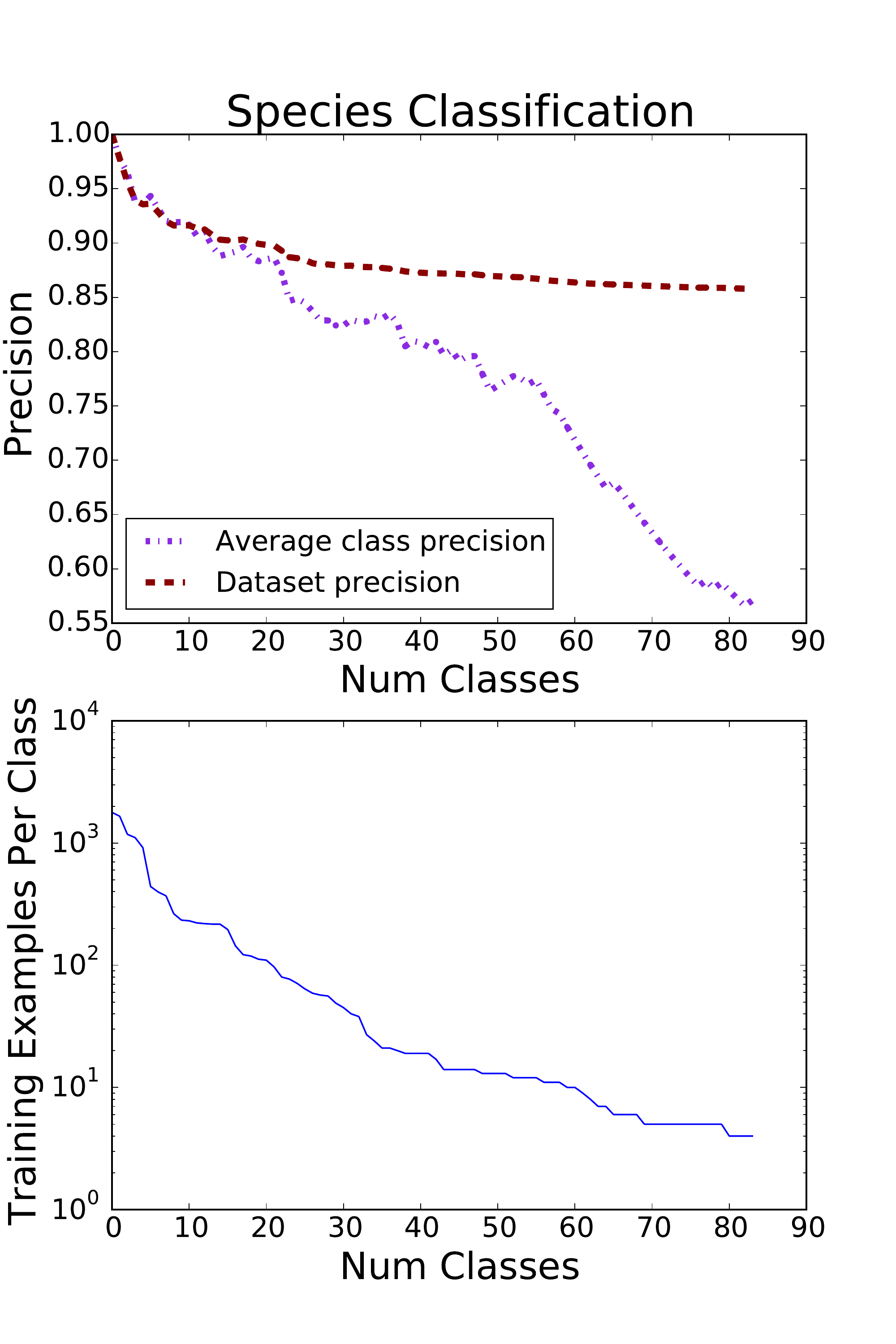} &
\includegraphics[width=.5\linewidth]{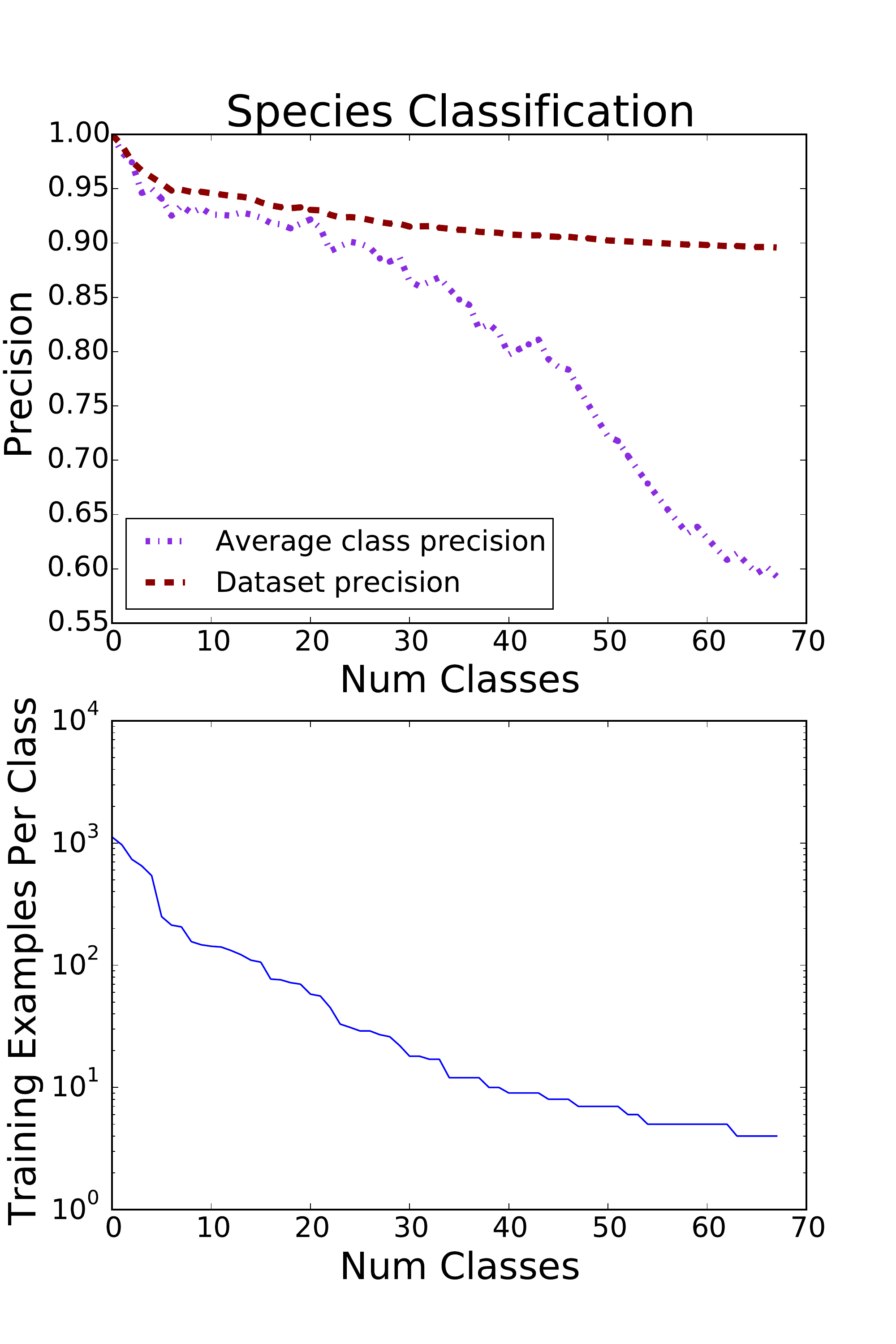}\\
(a) & (b)\\
\end{tabular}
\Caption{Comparison of tree species recognition results (for an increasing number of species) if using (a) ground truth positions of trees from the inventory and (b) positions from the detection pipeline. The bottom row plots training examples per class (log-scaled) versus number of classes, ordered by number of instances per class left to right.}
\label{fig:speciesclassifier}
\end{figure}
We compare the results of the end-to-end detection-recognition workflow to our earlier work~\citep{wegner2016}, where we use ground-truth tree positions for species recognition. We visualize dataset precision and average class precision for species recognition using ground-truth positions in Fig.~\ref{fig:speciesclassifier}(a) and those using detection positions in Fig.~\ref{fig:speciesclassifier}(b). We order precision values according to the number of classes, i.e. we first evaluate on the species with the highest number of instances, then on the first two with the highest number of instances and so on. The graphs in the bottom row of Fig.~\ref{fig:speciesclassifier} plot training examples per class versus number of classes. This together with average class precision plots in the top row help to judge the breaking point of the algorithm in terms of the minimum required number of training samples per tree species. It turns out that dataset-wide precision is always pretty good, while class average precision severely drops down once we have $<10$ training samples per class.
Species classification on detected tree locations (Fig.~\ref{fig:speciesclassifier}(b)) rather than ground truth locations (Fig.~\ref{fig:speciesclassifier}(a)) works just as well. In fact, average class precision is slightly higher on detected tree locations (e.g., 0.88 versus 0.83 for 30 different species). There are two main reasons for this: (i) detections center the tree in the image whereas ground-truth positions are often slightly misaligned with tree positions in the images, due to GPS inaccuracies in the reference data and inaccurate heading of the street-view panoramas; (ii) images at some positions of the reference data might actually not show a tree because it is (partially) occluded or has been entirely removed since the inventory took place in 2013.  
In general, our approach achieves $>0.80$ average class precision for the 40 most frequent tree species (Fig.~\ref{fig:speciesclassifier}(b)). However, average class precision already starts dropping significantly after the 20 most frequent tree species. This allows to make a rough estimate of the minimum number of training instances per species to achieve good performance. The $20^{th}$ most frequent species is the American Sweetgum with 155 occurrences in total, which leads to 109 instances for training (i.e., with a train, validation, test split of $70\%$, $20\%$, $10\%$). As a rule of thumb, we can say that a tree species needs $>100$ instances if using our setup of four images per tree and the limited amount of different species.  

\subsection{Tree change tracking results}\label{subsec:results_change tracking}
%
We split the 479 image pairs into a training, validation and test set for experiments. All input image patches are re-sampled to $256\times256$ pixels dimension to match the standard input requirement of the pre-trained VGG-16 network branches. Recall that we normalize all splits with mean and standard deviation of the training data set. Weights of feature extraction branches are initialized with pre-trained models and fine-tuned at train time while weights of the fully-connected layers are randomly initialized. Both branches of the Siamese CNN are initialized with the VGG-16 model that was trained for tree species classification. 
The ground truth data is split into 64\% training, 16\% validation and
20\% test data. Due to the limited data set size (479 image pairs in
total), 5-fold cross validation was carried out to reduce the bias of
individual train-test splits. All quantitative performance
measures are average numbers across all folds.
%
\begin{figure}[!ht]
\centering
\includegraphics[width=1.\linewidth]{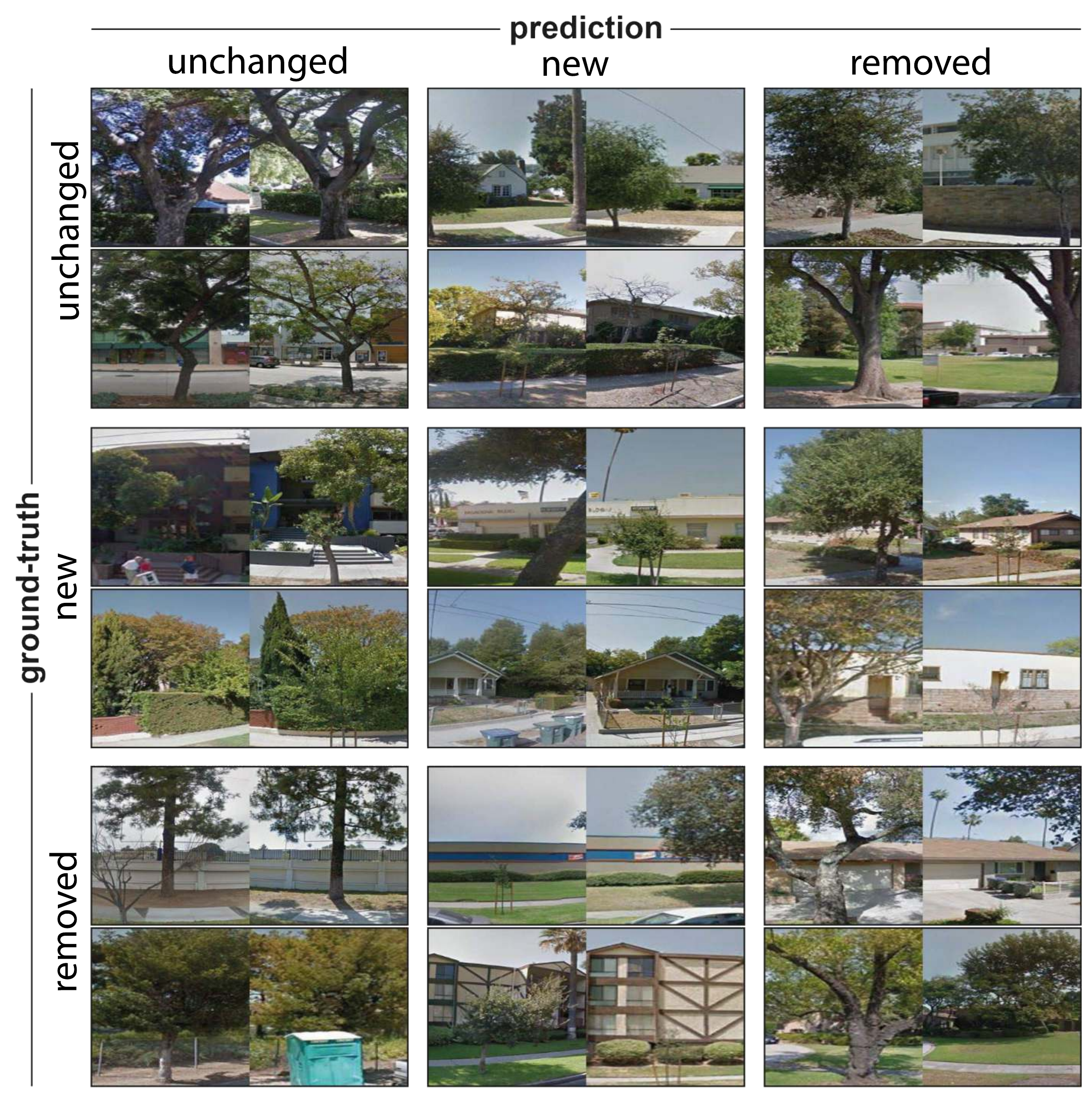}
\Caption{Confusion matrix with example images for the classification into \textit{unchanged}, \textit{new}, and \textit{removed} tree categories. Two example image pairs (left 2011, right 2016) are shown per case.}
\label{fig:confusionExamples}
\end{figure}
A minimum requirement for a (semi-)automated system in practice would be a binary classifier that distinguishes \textit{unchanged} from \textit{changed} trees, where a human operator could manually annotate the type of change for all trees classified changed. We test this setup first, before moving on to a multi-class model that directly predicts the type of change, too (i.e., \textit{unchanged}, \textit{new}, and \textit{removed} tree).
%
\begin{table}[]
\centering
\caption{Confusion matrix of the binary classification.}
\label{tab:convMatTreeSpeciesBINARY}
\begin{tabular}{ll | rr | ll}
                      &           & \multicolumn{2}{c}{predicted} \vline &       &        \\ 
                      &           & same    & changed   & total & recall \\ \hline
\multirow{2}{*}{\rotatebox[origin=c]{90}{true}} & same      & 183     & 17             & 200   & 0.915  \\
                      & changed       		& 21      & 258            & 279   & 0.925  \\ \hline
                      & total     		& 204     & 275          & 479   &        \\
                      & precision & 0.897   & 0.938      &       &       
\end{tabular}
\end{table}
%
\begin{table}[]
\centering
\caption{Confusion matrix of the multi-class classification}
\label{tab:convMatTreeSpeciesMulti}
\begin{tabular}{ll | rrr | ll}
                      &           & \multicolumn{3}{c}{predicted} \vline &       &        \\ 
                      &           & same    & new     & removed   & total & recall \\ \hline
\multirow{3}{*}{\rotatebox[origin=c]{90}{true}} & same      & 182     & 5      & 13         & 200   & 0.910  \\
                      & new       		& 4      & 119     & 8         & 131   & 0.908  \\
                      & removed   	& 10      & 2       & 136       & 148   & 0.919  \\ \hline
                      & total     		& 196     & 126     & 157       & 479   &        \\
                      & precision & 0.929   & 0.944   & 0.866     &       &       
\end{tabular}
\end{table}
For the binary classification we achieve 0.92 mean overall accuracy (mOA), whereas we get a slightly lower 0.91 mOA for the multi-class task. Confusion matrices for binary and multi-class results are shown in Tab.~\ref{tab:convMatTreeSpeciesBINARY} and Tab.~\ref{tab:convMatTreeSpeciesMulti}, respectively. We illustrate typical error cases of the multi-class task in Fig.~\ref{fig:confusionExamples}, where two example image pairs are randomly sampled and shown per failure mode in a confusion matrix-like scheme. 

A typical cause for errors is a slight change of heading direction between images of 2011 and 2016, which is due to inaccurate heading meta-data that comes with the street-view panoramas. A good example for this failure reason can be observed in the top row, right (ground-truth: unchanged, prediction: removed). While in both cases the tree is correctly shown in the center of the left (2011) image, the same tree is erroneously located on the right side of the right (2016) image leaving the image center empty. This causes the classifier to assign label \textit{removed}. An interesting case is shown in the middle row, right (ground-truth: new, prediction: removed). The left 2011 images show an old tree that has been replaced with a small, new tree in the right 2016 image. Practically speaking, both labels \textit{removed} and \textit{new} tree would be correct here which would call for a dedicated label \textit{old tree replaced with new tree}. However, since these cases where a small, new tree is planted at exactly the same location as an old tree before are very rare, we did not have enough training data to add them as a separate class. We decided to always label these situations as \textit{new} tree planted in the reference data, since the presence of a new tree is more important for future management than the previous state of the site. Another situation that is difficult to resolve for the classifier is if dense vegetation dominates both images (2011 and 2016) and individual trees can hardly be separated visually. This is shown in the middle row, left bottom (ground-truth: new, prediction: unchanged).

\section{Conclusions}\label{sec:conclusions}

We have presented a full tree detection and species recognition processing pipeline that facilitates 
the automated generation of tree inventories in urban scenarios. Experimental results are shown for the city of Pasadena and we plan to process more cities in the future to fully proof generalizability of our approach.
For the Pasadena dataset, it detects $>70\%$ of all trees correctly, potentially more if ground truth issues can be resolved. It also achieves $>80\%$ species recognition accuracy (i.e., average class precision) for the 40 most frequent tree species. A natural limit for machine learning-based approaches is the amount of available training data. 
As a rule of thumb, our pipeline needs at least 100 instances per tree species for training the CNN to achieve 
satisfying results. 
We also presented a tree change classification approach centered around a Siamese CNN that classifies the type of change 
per individual tree given two images of two different points in time. This approach delivers the correct result 
in $>90\%$ of the cases for three different change categories unchanged, removed, and new tree planted. 

We compare the system's performance with a recent study on manual collection of inventories of~\citet{roman2017}. The authors found that citizen scientists, i.e. volunteers that have between 1 and 3 years 
experience in urban forestry field work, occasionally miss trees (1.2\%) or count extra trees (1.0\%). Species are annotated 
consistently with expert arborists for 84.8\% of trees (\citet{cozad2005} report 80\% for this task). If we contrast these numbers with results of our automated system, we 
see that, unsurprisingly, detection performance falls short due to reasons analyzed in~\ref{subsubsec:detection_err_analysis} but species recognition performance is on par with citizen scientists. 
In general, detection, recognition, and change tracking results are very encouraging and motivate further research on more datasets, further cities, and in different geographic areas based on standard RGB images.


Beyond detection and species recognition it seems possible to estimate further parameters like the trunk diameter 
(which would allow to better estimate tree age and biomass) or the tree stress level. The latter would, for instance, allow to detect tree pests at an early stage and better direct resources, or to learn when local climate changes make a species less suitable for replanting.
In the future we plan to test our approach on more cities and with additional 
or different data. For example, only 60\% of the city of Zurich is covered with Google street-view panorama images (due 
to legal issues) but images of crowd-sourced street-view tools like \url{Mapillary.com}, or data collected by national mapping agencies could potentially fill in many gaps.

{
	  \bibliographystyle{elsarticle-harv}
		\bibliography{IJPRSTreeSpecies2017bib} 
}

\newpage
\appendix
\section{Projections between Images and geographic coordinates}\label{sec:projections}

We need to specify geometric projections between images and geographic coordinates to be able to turn tree detections in images into absolute real-world coordinates that will be useful in practice.   
To project tree detections in images to geographic coordinates and vice versa we have to specify mapping functions. We compute $\ell'=\mathcal{P}_v(\ell,c)$ to project a geographic position $\ell=(\mathrm{lat},\mathrm{lng})$ to image location $\ell'=(x,y)$ given camera parameters $c$, where index $v$ specifies the image modality, e.g., panorama at street-level or aerial image. In the following we formulate modality-specific projection functions.

A Web Mercator projection is used in Google maps for aerial images. Given geographic location $\ell=(\mathrm{lat},\mathrm{lng})$, a pixel location $\ell'=(x,y)$ is computed as $(x,y)=\mathcal{P}_{av}(\mathrm{lat},\mathrm{lng})$:
\begin{align}
 x &= 256 (2^{\mathrm{zoom}})
  \left(\mathrm{lng+\pi} \right)/2\pi\\
 y &= 256 (2^{\mathrm{zoom}}) \left(1/2 - \ln \left[\tan\left(\pi/4+\mathrm{lat}/2 \right)\right] / 2\pi \right)
 \label{eq:aerial_geo2pix}
\end{align}  
where $\mathrm{zoom}$ defines the image resolution.
In order to project tree detections in aerial images to geographic coordinates we need the inverse function, too. Using simple algebraic manipulation of Eq.~\ref{eq:aerial_geo2pix}, the inverse function $(\mathrm{lat},\mathrm{lng})=\mathcal{P}_{av}^{-1}(x,y)$ can be computed as:
\begin{equation}
\begin{split}
 \mathrm{lng} &= \frac{\pi x}{128 (2^{\mathrm{zoom}})} - \pi \\
 \mathrm{lat} &= 2 \tan^{-1} \left( \exp\left( \pi - \frac{y \pi}{128 (2^{\mathrm{zoom}})} \right) \right) - \frac{\pi}{4}
 \label{eq:aerial_pix2geo}
\end{split}
\end{equation}

Note that our annotations at both train and test time represent trees as point-wise latitude/longitude locations rather than bounding boxes. To build a training set, we use fixed size boxes around points $(x_j,y_j)=\mathcal{P}_{av}(\mathrm{lat}_j,\mathrm{lng}_j)$ for each ground truth tree $(\mathrm{lat}_j,\mathrm{lng}_j)$. In practice, we use $100 \times 100$ boxes, which correspond to regions of $\approx 12 \times 12$ meters. In this scenario, boxes should be interpreted as boxes for feature extraction rather than bounding boxes around trees. A test time detection of a box centered at $(x,y)$ results in a geographic prediction of $(\mathrm{lat},\mathrm{lng})=\mathcal{P}_{av}^{-1}(x,y)$. Note that map images are pixel-wise aligned with aerial view images and thus subject to the same projection functions. Maps contain rasterized 2D graphics of streets, buildings, parks, etc. and will be useful to formulate probabilistic priors within the CRF framework as we will see in Sec.~\ref{sec:detection-multiview-prob}. 
\begin{figure}[t]
\centering
\begin{tabular}{cc}
\includegraphics[width=0.40\linewidth]{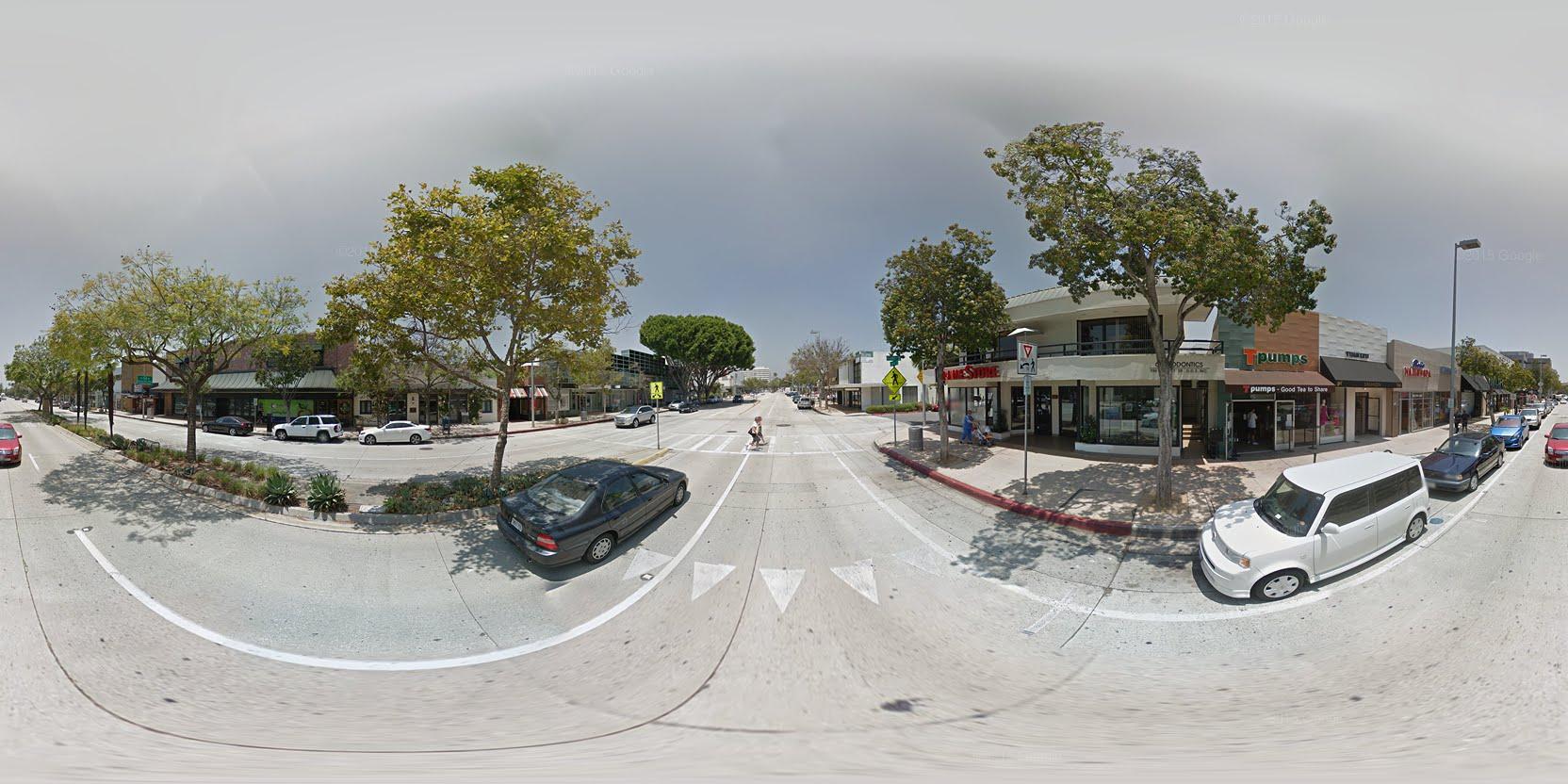} &
\includegraphics[width=0.52\linewidth]{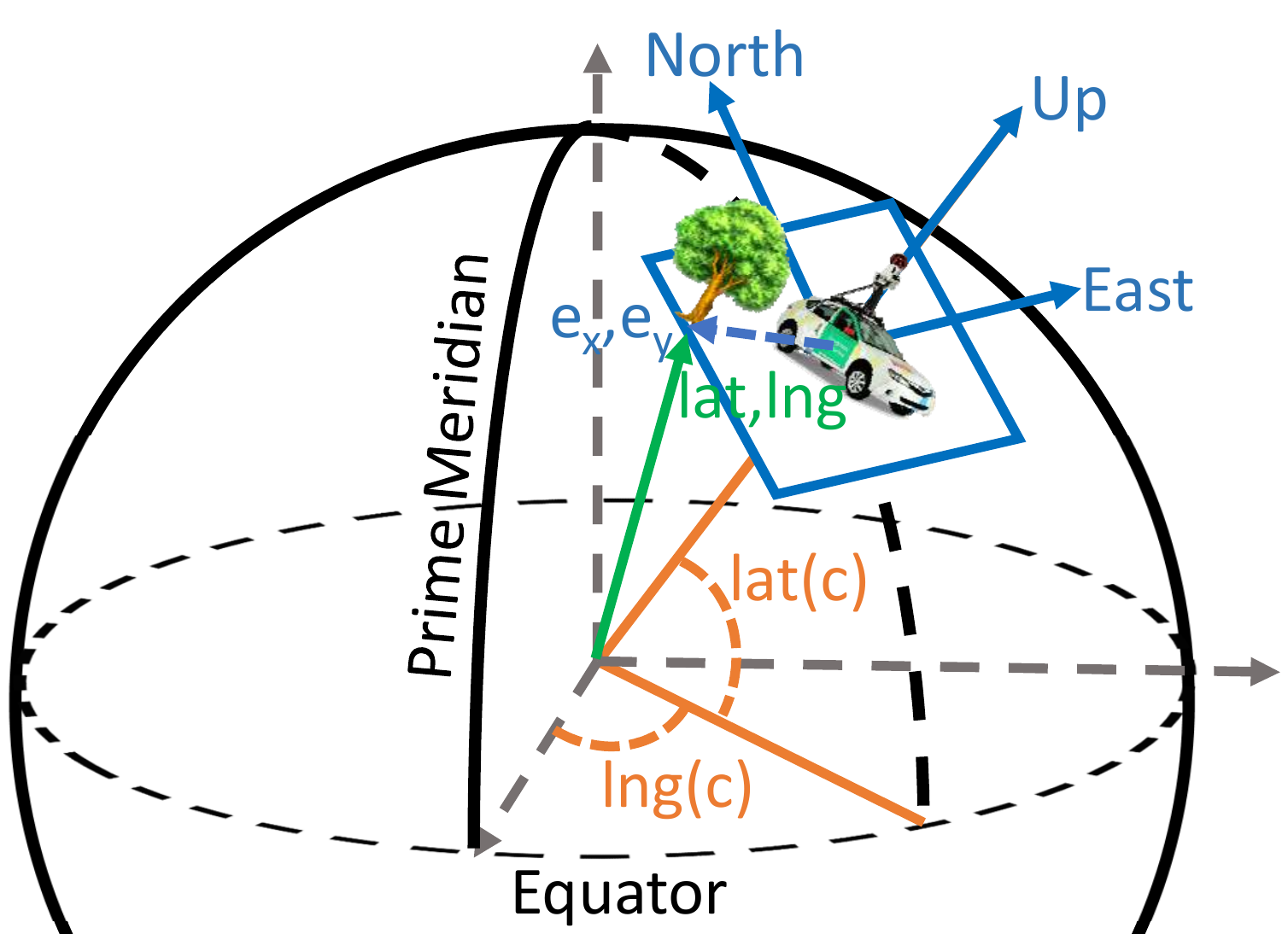} \\
(a) & (b)\\
\end{tabular}
\Caption{\textbf{(a)} $360 \degree$ street view panorama image from Google Maps. \textbf{(b)} Geographic positioning of a street view camera at geographic coordinate (lat,lng), where each object is represented in ENU coordinates at a plane that is locally tangent to the earth's surface.}
\label{fig:pano_tree_position}
\end{figure}

Each Google street view image captures a full $360 \degree$ panorama (Fig.~\ref{fig:pano_tree_position}(a)) and is an equidistant cylindrical projection of the environment as captured by a camera mounted on top of the Google street view car. The car is equipped with GPS and an inertial measurement unit to record its camera position $c$, which includes the camera's geographic coordinates $\mathrm{lat}(c)$, $\mathrm{lng}(c)$, and the car's heading $\mathrm{yaw}(c)$ (measured as clockwise angle from north). On urban roads, Google street view images are typically spaced $\approx 15$ m apart.

We make the simplifying assumption of locally flat terrain to estimate geographic object coordinates from individual street view panoramas. Each object is represented in local East, North, Up (ENU) coordinates with respect to the geographic position of the camera (see Figure~\ref{fig:pano_tree_position}(b)), where the $x$-axis points east, the $y$-axis points north, and the $z$-axis points up. The ENU position $(e_x,e_y,e_z)$ of a tree at $(\mathrm{lat},\mathrm{lng})$ then is
\begin{equation}
(e_x,e_y,e_z) = \bigl( R \cos[\mathrm{lat}(c)]\sin[\mathrm{lng}-\mathrm{lng}(c)], R\sin[\mathrm{lat}-\mathrm{lat}(c)], -h \bigr)
\end{equation}

with $h$ the height of the camera above ground and earth radius $R$. A tree's location on the ground plane is then at a clockwise angle of $\arctan(e_x, e_y)$ from north (with tilt $\arctan(-h, z)$) with distance $z=\sqrt{e_x^2+e_y^2}$ from the camera (Fig.~\ref{fig:pano_tree_position}(b)). 

With the camera's heading, ENU coordinates can be turned into cylindrical coordinates. We then get image projection $(x,y)=\mathcal{P}_{sv}(\mathrm{lat},\mathrm{lng},c)$ to obtain image coordinates $\ell'=(x,y)$ as
\begin{equation}
\begin{split}
 x =& \left(\pi + \arctan(e_x, e_y) - \mathrm{yaw}(c)\right)W/{2\pi}\\
 y =& \left(\pi/2 - \arctan(-h, z) \right) H/{\pi}
 \label{eq:streetview_geo2pix}
\end{split}
\end{equation}
where W and H represent width and height of the panorama.   
Since our annotations at train and test time represent trees as point-wise latitude/longitude locations rather than bounding boxes, we compute each training bounding box as the box that a $8 \times 12$ meter object would occupy (using Eq~\ref{eq:streetview_geo2pix}).


\end{document}